\documentclass{article}

 \usepackage[main, final]{neurips_2026}

\usepackage[utf8]{inputenc} 
\usepackage[T1]{fontenc}    
\usepackage{hyperref}       
\usepackage{url}            
\usepackage{booktabs}       
\usepackage{amsfonts}       
\usepackage{nicefrac}       
\usepackage{microtype}      
\usepackage[table,dvipsnames]{xcolor}         

\hypersetup{
colorlinks = true,
linkcolor = RoyalBlue,
anchorcolor = blue,
citecolor = RoyalBlue,
filecolor = cyan,
menucolor = ForestGreen,
runcolor = cyan,
urlcolor = ForestGreen
}

\usepackage{wrapfig}
\usepackage{graphicx}

\usepackage{multicol}

\usepackage{multirow}
\usepackage{tabu}

\usepackage{pifont}
\usepackage{adjustbox}
\usepackage{amsmath}
\usepackage{xspace}

\usepackage[capitalize,noabbrev]{cleveref}
\usepackage{threeparttable}

\usepackage{tcolorbox}

\usepackage{amsthm}
\theoremstyle{plain}

\theoremstyle{definition}

\theoremstyle{remark}

\usepackage{algorithmic}
\usepackage[vlined,commentsnumbered,linesnumbered,ruled]{algorithm2e}

\usepackage{diagbox}

\newcommand{\LATER}[1]{}

\usepackage{dsfont}

\usepackage{graphicx} 

\usepackage{tikz}
\usetikzlibrary{arrows.meta}

\def\secref#1{\S\ref{sec:#1}}
\def\seclabel#1{\label{sec:#1}}

\usepackage[table]{xcolor}
\definecolor{GreedyBlue}{HTML}{003366}   
\definecolor{SampleGreen}{HTML}{006400}  
\definecolor{paleviolet}{HTML}{F4EEFF}
\newcommand{\greedy}[1]{\textbf{\textcolor{GreedyBlue}{#1}}}     
\newcommand{\sample}[1]{\textbf{\textcolor{SampleGreen}{#1}}}  

\title{Consolidating Rewarded Perturbations for\\ LLM Post-Training}

\author{
    Zheyu Zhang$^{*}$, Shuo Yang$^{*}$, Gjergji Kasneci \vspace{0.2cm}\\
    Technical University of Munich \\
    Munich Center for Machine Learning (MCML) \\
    {\small \tt \{name.surname\}@tum.de} \\
}

\begin{document}

\maketitle

\begingroup
\renewcommand{\thefootnote}{\fnsymbol{footnote}}
\footnotetext[1]{Equal contribution.}
\endgroup

\begin{abstract}
Post-training of language models is commonly framed as a sample-score-update loop implemented by gradient descent. A recent line of work, exemplified by RandOpt, relocates this loop to weight space, sampling Gaussian perturbations around a pretrained model and ensembling the top-$K$ rewarded specialists at inference. While competitive with PPO and GRPO under matched training compute, this prediction-level ensemble incurs $K$ forward passes per test example and does not extend cleanly to free-form generation. We ask whether the rewarded population can instead be folded into a single deployable model, replacing the inference-time ensemble with one consolidated update.
A split-half analysis over 25 model-task pairs reveals reproducible low-rank structure in every case.
We turn this geometry into \textbf{CoRP} (\textbf{Co}nsolidating \textbf{R}ewarded \textbf{P}erturbations), a gradient-free operator that combines reward-weighted aggregation, compatibility-aware reweighting, and a held-out validation gate, with no gradient flowing through the language model.
Across five language models from $0.5$B to $8$B and five tasks covering math, code, and creative writing, CoRP improves the base model by $8.1$ points on average. Using one tenth of RandOpt's perturbation budget, CoRP exceeds single-inference RandOpt by $6.5$ points and recovers more than half of the gain of the 50-pass majority-vote ensemble, at one forward pass per test example.
\end{abstract}

\section{Introduction}

Post-training is the alignment stage, where a broadly capable language model is adapted to produce responses preferred by humans while maintaining or improving performance, especially on reasoning-intensive tasks. Two families dominate current practice. \emph{Reinforcement learning with verifiable rewards (RLVR)}~\citep{ouyang2022rlhf, schulman2017ppo, shao2024grpo, lambert2025tulu} drives much of the recent progress on reasoning models~\citep{deepseek2025r1, openai2024openaio1card, kimi2025k15}, and \emph{on-policy distillation (OPD)}~\citep{agarwal2024onpolicykd} trains the model on its own trajectories under a stronger teacher and now appears as a core stage in pipelines like Qwen3 and MiMo~\citep{yang2025qwen3, xiao2026mimo}. Different as the two families look, both follow the same pattern: sample from the model, derive a per-trajectory or per-token learning signal, and update the weights to amplify capabilities that pretraining placed within reach.

\begin{figure}[t]
\centering
\begin{adjustbox}{width=\linewidth,center}
\begin{tikzpicture}[
    every node/.style={font=\small, text=black!88},
    box/.style={draw=black!35, rounded corners=2pt, line width=0.55pt, minimum height=0.9cm, minimum width=1.66cm, align=center, font=\footnotesize, inner sep=2.5pt, fill=black!1},
    final/.style={draw=NavyBlue!55, rounded corners=2pt, line width=0.65pt, minimum height=0.9cm, minimum width=1.66cm, align=center, font=\footnotesize\bfseries, text=NavyBlue!85!black, fill=NavyBlue!7, inner sep=2.5pt},
    finalg/.style={draw=ForestGreen!50!black, rounded corners=2pt, line width=0.65pt, minimum height=0.9cm, minimum width=1.66cm, align=center, font=\footnotesize\bfseries, text=ForestGreen!45!black, fill=ForestGreen!6, inner sep=2.5pt},
    arrow/.style={-{Latex[length=2.4mm,width=1.7mm]}, line width=0.9pt, draw=black!65},
    rowlabel/.style={font=\footnotesize\bfseries, align=center, text width=1.0cm, text=black!65},
    methodlabel/.style={font=\scriptsize, align=center, text=black!72, text width=1.7cm}
]
\node[rowlabel] (lab1) at (-6.0, 0.40) {Output\\space};
\node[box] (m1) at (-4.10, 0.52) {Base\\model $\theta_0$};
\node[box] (s1) at (-1.00, 0.52) {Sampled\\trajectories $\tau$};
\node[box] (r1) at (2.05, 0.52) {Reward /\\teacher signal};
\node[final] (u1) at (5.10, 0.52) {Updated\\$\theta'$};
\draw[arrow] (m1) -- node[above, font=\scriptsize] {sample} (s1);
\draw[arrow] (s1) -- node[above, font=\scriptsize] {score} (r1);
\draw[arrow] (r1) -- node[above, font=\scriptsize] {gradient} (u1);
\node[methodlabel] at (6.95, 0.52) {RLVR, OPD};
\node[rowlabel] (lab2) at (-6.0, -0.75) {Weight\\space};
\node[box] (m2) at (-4.10, -0.75) {Base\\weights $\theta_0$};
\node[box] (s2) at (-1.00, -0.75) {Perturbations\\$\theta_0 + \Delta_i$};
\node[box] (r2) at (2.05, -0.75) {Support-set\\rewards $r_i$};
\node[finalg] (u2) at (5.10, -0.75) {Updated\\$\hat\theta$};
\draw[arrow] (m2) -- node[above, font=\scriptsize] {sample} (s2);
\draw[arrow] (s2) -- node[above, font=\scriptsize] {score} (r2);
\draw[arrow] (r2) -- node[above, font=\scriptsize] {select} (u2);
\node[methodlabel] at (6.95, -0.75) {RandOpt: vote\\\textbf{CoRP: merge}};
\end{tikzpicture}
\end{adjustbox}
\caption{Two routes for self-elicitation post-training. 
Output-space methods sample trajectories and fold reward or teacher feedback back through gradients. 
Weight-space methods sample perturbations around $\theta_0$ and fold rewarded variation back by aggregation. 
CoRP follows the weight-space route and returns a single deployable model instead of an inference-time ensemble.}
\vspace{-0.4cm}
\label{fig:pipeline}
\end{figure}
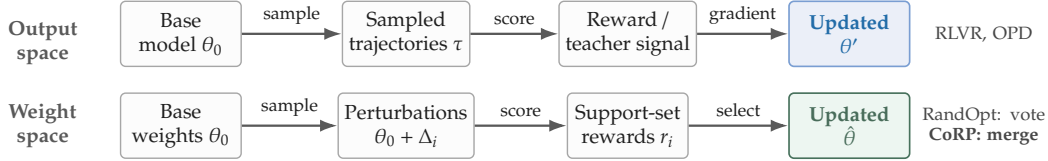

A growing body of evidence suggests that what these updates accomplish is mainly to activate capabilities pretraining already made accessible. Chain-of-thought prompting elicits multi-step reasoning from models that were never trained on intermediate traces~\citep{wei2022cot, kojima2022zeroshot}. RLVR continues to improve a base model even when its reward is partially incorrect or spurious, which motivates the hypothesis that the gradient update may operate beyond direct reward maximization~\citep{shao2025spurious}. Mechanistic studies of reasoning fine-tuning reach a similar conclusion. They find that the update repurposes directions that are already active in the base model rather than installing new ones~\citep{ward2025reasoning, liu2025understanding, gandhi2025cognitive,li2026learning}. Taken together, these results suggest that post-training often exposes and stabilizes capabilities the pretraining step had already placed within reach.
 
As our baseline, \citet{gan2026neuralthickets} push this view from output space to weight space. The neighborhood of a contemporary pretrained model is dense with task specialists, and a single round of rewarded Gaussian perturbations is enough to expose them. RandOpt evaluates each perturbation on a small support set, keeps the top-$K$ specialists, and votes their predictions at inference. The resulting ensemble matches or exceeds common post-training methods like PPO and GRPO at equal training compute, with no gradient through the language model. Where output-space recipes sample what the model can say, RandOpt samples what the model could be (see \Cref{fig:pipeline}).

This is effective at training but expensive at inference. Good performance needs $K \approx 50$, which means $50$ forward passes per test example, and the prediction-level vote has to be redesigned for tasks whose outputs are not categorical. The natural fix is to consolidate the rewarded population into a single weight update. Reward-weighted averaging is the obvious first attempt, but rewarded perturbations behave as nearly orthogonal specialists, so naive averaging cancels rather than combines. Existing model-merging tools~\citep{wortsman2022soups, frankle2020linearmode, ilharco2023task, yadav2023ties, yu2024dare, matena2022fisher} assume independently fine-tuned checkpoints. Since random rewarded perturbations around a single base model are not fine-tuned checkpoints, their compatibility must be evaluated directly.
 
Our study begins with a measurement: Across 25 model-task pairs spanning five families and five tasks, a reward-driven random search around one base model concentrates its useful variance on a reproducible low-rank subspace in every single case we examine. A reproducible mean direction, by contrast, exists in only 40 percent of the cases. These results indicate the presence of shared structure, which is distributed across a family of compatible directions rather than a single broad mean.
 
Building on this, we introduce \emph{Consolidating Rewarded Perturbations} (CoRP\footnote{Our code is available at \url{https://github.com/oooranz/CoRP}.}), a gradient-free operator that turns a rewarded perturbation population into one deployable update. CoRP first forms a reward-weighted candidate direction. It then reweights each perturbation by its alignment with that direction and its orthogonal dispersion, suppressing high-reward perturbations that would inject incompatible structure. A held-out validation gate commits a candidate only when its gain is stable on examples that did not contribute to its construction~\citep{stone1974cv, chernozhukov2018doubleml}, and lets the method abstain when no candidate is reliable. Across five language models from $0.5$B to $8$B parameters and five tasks, CoRP uses $500$ rewarded perturbations, one tenth of RandOpt's budget, and recovers more than half of its $K{=}50$ ensemble gain at one forward pass per test example.

\section{Background}
\seclabel{sec:background}

\subsection{Rewarded Perturbations}

\citet{gan2026neuralthickets} treat post-training as a problem of sampling and filtering in weight space. Let $\theta_0 \in \mathbb{R}^d$ denote the pretrained weights of a language model. Given a finite set of noise scales $\Sigma \subset \mathbb{R}_{>0}$ and a sample size $N$, RandOpt draws $N$ Gaussian perturbations of $\theta_0$,
\begin{equation}
\Delta_i \;=\; \sigma_i\, \xi_i,
\qquad \xi_i \sim \mathcal{N}(0, I_d),
\qquad \sigma_i \in \Sigma,
\qquad i = 1, \dots, N,
\label{eq:randopt-sample}
\end{equation}
where each $\sigma_i$ is sampled uniformly from $\Sigma$ and each $\xi_i$ is an independent isotropic Gaussian direction in $\mathbb{R}^d$. RandOpt then scores each perturbed model on a small support set $D_{\mathrm{sup}}$ of task examples,
\begin{equation}
r_i \;=\; \frac{1}{|D_{\mathrm{sup}}|}\sum_{x \in D_{\mathrm{sup}}} R(\theta_0 + \Delta_i;\; x),
\label{eq:randopt-reward}
\end{equation}
where $R(\theta;\,x) \in \mathbb{R}$ is the task reward of the model with weights $\theta$ on input $x$. We refer to the resulting set $\mathcal{P} = \{(\Delta_i, r_i)\}_{i=1}^N$ a \emph{rewarded perturbation population}. It is the basic object our method operates on.
 
\subsection{Prediction-Level Ensembling}

RandOpt predicts by retaining the top-$K$ entries of a rewarded population and voting over the corresponding perturbed models,
\begin{equation}
E_K = \operatorname{TopK}\!\big(\{r_i\}_{i=1}^N\big),
\quad
\hat y(x') = \operatorname{MajVote}\!\big(\{f(\theta_0 + \Delta_i;\, x') : i \in E_K\}\big),
\label{eq:randopt-ensemble}
\end{equation}
where $f(\theta;\,x)$ denotes the model output at input $x$ under weights $\theta$. The ensemble works because the neighborhood of a well-pretrained $\theta_0$ contains many task-improving perturbations, and because those perturbations behave as specialists whose strengths increasingly diverge as scale grows~\citep{gan2026neuralthickets}. The same two facts make a single update hard. It has to preserve what the rewarded population shares while ignoring directions that are individually strong but mutually incompatible.

\section{Do Rewarded Perturbations Share a Common Structure?}
\seclabel{sec:characterization}

A rewarded population is dense and diverse. Whether it also shares a reproducible structure is a separate question, and one the prediction-level ensemble does not need to answer. A pairwise test cannot answer it either. Independent Gaussian directions in a billion-dimensional parameter space are nearly orthogonal whether they are rewarded or not~\citep{vershynin2018hdp}, so any shared signal has to be read at the population level. We read it from a split-half diagnostic in the spirit of classical reliability analysis~\citep{brown1910split, spearman1910split}.

\begin{wrapfigure}{r}{0.5\columnwidth} 
\vspace{-0.6cm}
\includegraphics[width=0.5\columnwidth]{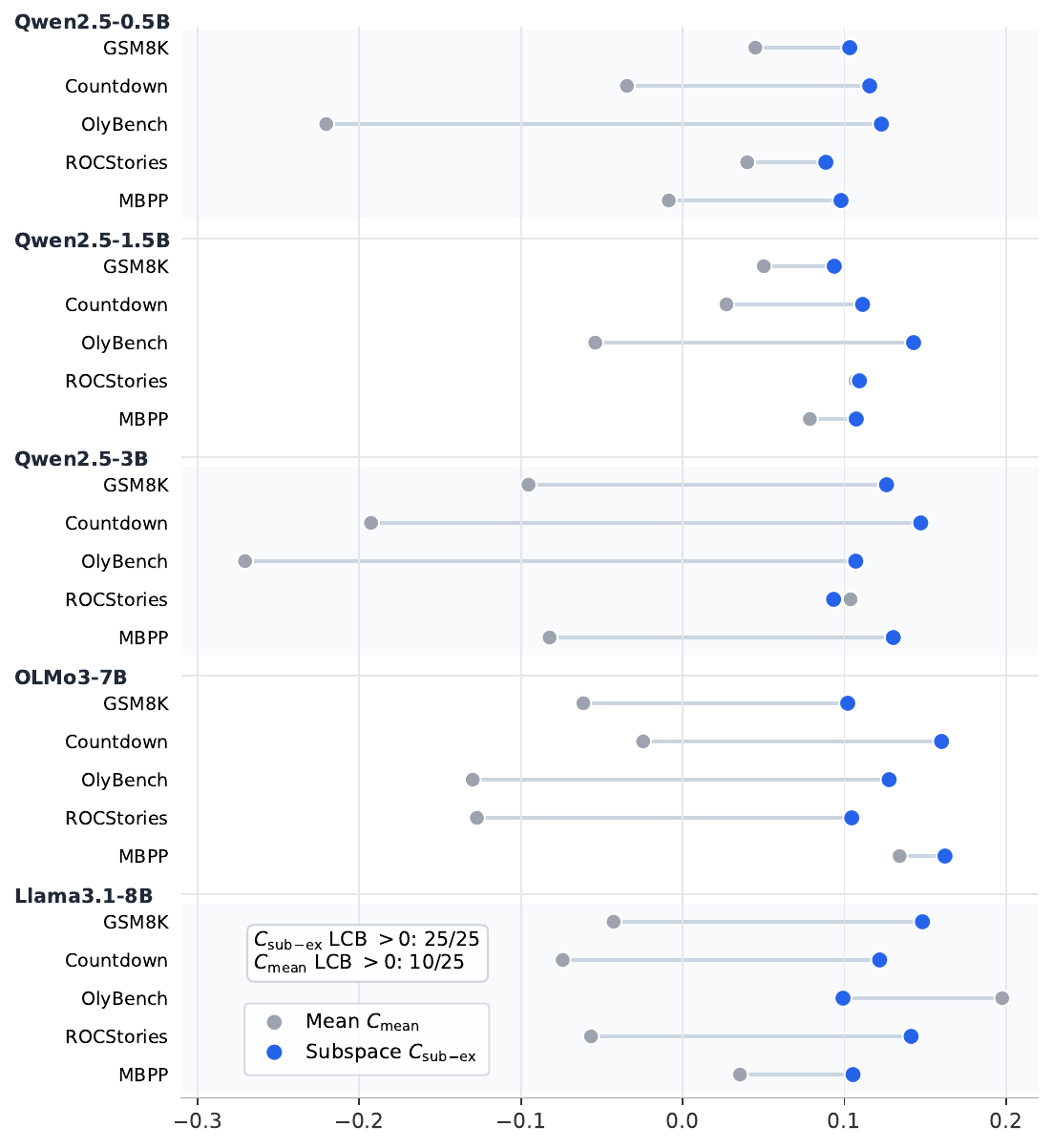}
\vspace{-0.6cm}
\caption{Split-half statistics at $M{=}50$. Each row is one model-task pair. Gray points are the lower confidence bound of the mean statistic $C_{\mathrm{mean}}$, blue points the lower confidence bound of the subspace statistic $C_{\mathrm{sub\text{-}ex}}$. The dashed line marks zero.}
\label{fig:characterization}
\end{wrapfigure}
 
For each rewarded population we project the perturbations onto a fixed low-dimensional coordinate sketch $z_i \in \mathbb{R}^m$ with $m \ll d$. We rank the population by reward, partition the top-$M$ indices into two disjoint halves $S_A$ and $S_B$ of equal size, and repeat the partition over 20 random splits. On each half we form a reward-weighted mean of the sketched perturbations, $\mu_A = \sum_{i \in S_A} w_i^{(1)} z_i$ and $\mu_B = \sum_{i \in S_B} w_i^{(1)} z_i$, with the reward-tilted weights $w_i^{(1)}$ defined in Eq.~\ref{eq:m1} below. The first statistic compares the two means,
\begin{equation}
C_{\mathrm{mean}}(M)
\;=\;
\mathbb{E}_{\mathrm{split}}\!\big[\cos(\mu_A,\,\mu_B)\big],
\label{eq:mean-consensus}
\end{equation}
where the expectation is taken over random splits. The second statistic compares the top-$r$ principal subspaces of the two weighted clouds, with the random-baseline overlap $r/m$ between two random $r$-dimensional subspaces of $\mathbb{R}^m$ subtracted away,
\begin{equation}
C_{\mathrm{sub\text{-}ex}}(M, r)
\;=\;
\mathbb{E}_{\mathrm{split}}\!\big[\,\tfrac{1}{r}\,\big\|U_A^\top U_B\big\|_F^2\,\big]
\;-\;
\tfrac{r}{m},
\label{eq:subspace-excess}
\end{equation}
where $U_A, U_B \in \mathbb{R}^{m \times r}$ are orthonormal bases of the top-$r$ principal subspaces of the weighted point clouds on the two halves and $r$ is fixed at a value reported in the appendix. We call a statistic \emph{stable} when its nonparametric 95 percent lower confidence bound over the random splits lies above zero.
 
We run the diagnostic on 25 model-task pairs across five instruction-tuned models, Qwen2.5-0.5B-Instruct, Qwen2.5-1.5B-Instruct, Qwen2.5-3B-Instruct, OLMo3-7B-Instruct, and Llama-3.1-8B-Instruct, and across five tasks covering arithmetic search (Countdown), grade-school math (GSM8K), olympiad-style reasoning (OlympiadBench), creative writing (ROCStories), and program synthesis (MBPP). \Cref{fig:characterization} reports the result. The subspace-excess statistic is positive with 95 percent confidence on 25 of 25 pairs. The mean-consensus statistic clears the same bar on only 10. Rewarded perturbations therefore share a reproducible signal in every case we examine, but in most cases that signal lives in a low-rank family of compatible directions rather than in any single mean direction.
 
Two consequences follow. First, reward-weighted averaging cannot be the answer on its own, because the reward-weighted mean is only reproducible on a minority of pairs. Second, the shared signal nonetheless exists in every pair, distributed across a low-rank family of directions rather than collapsed onto one. A consolidation operator therefore needs to do two things at once: pick a working direction from the rewarded population, and handle perturbations whose strengths lie outside it.

\section{Consolidating Rewarded Perturbations}
\seclabel{method}
 
Reward-weighted averaging is the natural first attempt at consolidation, but it conflates usefulness with compatibility. A perturbation can earn a high reward on the support set and still point in a direction that the rest of the rewarded population does not share. Averaging then writes both the useful component of the perturbation and its incompatible component into the model, and the incompatible component is what cancels.
 
CoRP separates these two questions and answers them in sequence. A first pass uses reward alone to propose a direction. A second pass then asks which perturbations support the proposal and which add mass orthogonal to it, and reweights accordingly. A held-out validation gate commits the resulting candidate only when it continues to help on examples not used to construct it, and abstains when no candidate is reliable. No gradient flows through the language model at any step.

\subsection{From Reward to Compatibility}
\seclabel{operator}

Let $\mathcal{P} = \{(\Delta_i, r_i)\}_{i=1}^N$ be the rewarded perturbation population defined in Eqs.~\ref{eq:randopt-sample}--\ref{eq:randopt-reward}. CoRP builds a single weight update from this population by assigning weights to the sampled perturbations and normalizing the resulting direction.

The first pass retains the top-$q$ reward quantile,
\begin{equation}
E_q \;=\; \{\,i : r_i \ge \tau_q\,\},
\qquad
\text{where } \tau_q \text{ is the $q$-th quantile of } \{r_i\}_{i=1}^N,
\label{eq:elite}
\end{equation}
and forms reward-tilted weights and a provisional direction,
\begin{equation}
w_i^{(1)}
\;=\;
\frac{\exp(\beta\, r_i)\,\mathbf{1}[i \in E_q]}{\sum_{j} \exp(\beta\, r_j)\,\mathbf{1}[j \in E_q]},
\qquad
m^{(1)} \;=\; \sum_{i=1}^N w_i^{(1)}\, \Delta_i,
\label{eq:m1}
\end{equation}
where $\beta > 0$ is an inverse temperature and $\mathbf{1}[\cdot]$ is the indicator function. The provisional direction $m^{(1)} \in \mathbb{R}^d$ is the reward-weighted mean of the elite perturbations. This first pass follows the weighted-sample logic used in reward-weighted regression, advantage-weighted regression, the cross-entropy method, and evolution strategies~\citep{peters2007rwr, peng2019awr, rubinstein1999cem, mannor2003cem, salimans2017es}. It is not yet a deployable update. Reward selects perturbations that help the task, but it does not say whether those perturbations can be written into the same model.

The second pass measures compatibility with the proposal. For each elite perturbation, CoRP computes its alignment with $m^{(1)}$ and its orthogonal dispersion against $m^{(1)}$,
\begin{equation}
a_i \;=\; \cos\!\big(\Delta_i,\, m^{(1)}\big),
\qquad
d_i \;=\; \big\|\Delta_i - \Pi_{m^{(1)}}\Delta_i\big\|^2,
\label{eq:align-disp}
\end{equation}
where $\Pi_{m^{(1)}}\Delta_i$ is the projection of $\Delta_i$ onto the line spanned by $m^{(1)}$. The alignment $a_i \in [-1, 1]$ measures how strongly a perturbation supports the proposal, and the dispersion $d_i \ge 0$ measures how much of it lies outside. CoRP then combines reward and compatibility into the second-pass weights,
\begin{equation}
w_i^{(2)}
\;=\;
\frac{\exp\!\left(\beta\, r_i + \gamma_a\, z(a_i) - \gamma_d\, z(d_i)\right)\,\mathbf{1}[i \in E_q]}{\sum_{j} \exp\!\left(\beta\, r_j + \gamma_a\, z(a_j) - \gamma_d\, z(d_j)\right)\,\mathbf{1}[j \in E_q]},
\label{eq:w2}
\end{equation}
where $\gamma_a, \gamma_d \ge 0$ control the strength of the compatibility terms and $z(\cdot)$ standardizes a statistic to zero mean and unit variance over $E_q$. The consolidated direction and update are
\begin{equation}
\bar{\Delta} \;=\; \sum_{i=1}^N w_i^{(2)}\, \Delta_i,
\qquad
\hat{\theta} \;=\; \theta_0 + \eta\, \bar{\Delta} \big/ \|\bar{\Delta}\|,
\label{eq:CoRP-step}
\end{equation}
with deployed step size $\eta > 0$. The three terms in the exponent of Eq.~\ref{eq:w2} capture, in order, usefulness, support for the proposal, and resistance to orthogonal mass. The same principle of measuring compatibility before aggregation appears in parameter-space merging of independently fine-tuned checkpoints~\citep{matena2022fisher, yadav2023ties, yu2024dare}. CoRP applies it to a different source family, random rewarded perturbations around one base model, where compatibility cannot be read from sign, magnitude, or curvature of fitted checkpoints and has to be measured against a population proposal.

\subsection{Validating the Consolidated Update}
\seclabel{crossfit}

The same support examples that weight the perturbations can also overstate the value of the resulting update. CoRP separates proposal generation from proposal validation, in the spirit of nested cross-validation and cross-fit estimation~\citep{stone1974cv, chernozhukov2018doubleml}. We split $D_{\mathrm{sup}}$ into three disjoint folds $A$, $B$, and $P$. For each configuration in a small grid of $(q, \beta, \eta)$, CoRP forms a candidate $\hat{\Delta}_{q,\beta}$ from rewards computed only on $A$, following Eqs.~\ref{eq:elite}--\ref{eq:CoRP-step}, and ranks candidates on $A$ by a constructive score
\begin{equation}
s^A(q, \beta, \eta)
\;=\;
\frac{|F^A(q, \beta, \eta)| - \lambda\, |G^A(q, \beta, \eta)|}{|A|},
\label{eq:constructive}
\end{equation}
where $F^A$ is the set of examples in $A$ that candidate $\theta_0 + \eta\, \hat{\Delta}_{q,\beta}$ answers correctly while base model $\theta_0$ does not, $G^A$ is the symmetric set of regressions, and $\lambda > 0$ penalizes regressions. A candidate passes the gate only if its constructive score on $B$ is positive and the lower confidence bound of its accuracy change on $B$ is positive. The probe set $P$ then calibrates the deployed step size on a fixed multiplier grid $\{\alpha_j\}$, retaining the largest $\alpha_j$ such that $\theta_0 + \alpha_j\, \eta\, \hat{\Delta}_{q,\beta}$ has a positive lower confidence bound on $\Delta\mathrm{Acc}_P$. If no candidate passes the gate or no multiplier passes the probe, CoRP abstains and returns $\theta_0$.

\subsection{Iterating Around an Accepted Update}
\seclabel{refine}

An accepted update changes the local neighborhood, and the rewarded population around the new center can carry useful structure that the original sampling did not reach. CoRP therefore iterates. The candidate construction, the held-out gate, and the probe calibration of the previous two subsections all remain in place. Only the sampling distribution changes between iterations.

Let $\theta_t$ denote the current center, with $\theta_0$ the pretrained model and $\theta_1$ the first accepted update. CoRP draws $N_{\mathrm{loc}}$ local perturbations
\begin{equation}
\Delta_j^{\mathrm{loc}} \;\sim\; \mathcal{N}\!\big(0,\, \rho_t^2\, \Sigma_t\big),
\qquad j = 1, \dots, N_{\mathrm{loc}},
\label{eq:local}
\end{equation}
where $\rho_t > 0$ is the local search scale and $\Sigma_t$ is a covariance whose low-rank component concentrates exploration in directions where the previous rewarded population varied, with an isotropic floor preserving coverage in directions the population did not emphasize. The exact form of $\Sigma_t$ follows covariance-adaptive black-box search~\citep{hansen2016cmaes, maheswaranathan2018guidedes, choromanski2019asebo} and is given in the appendix. The covariance shapes the proposal distribution only and does not appear in the deployed update. The next center $\theta_{t+1}$ is committed only when the validation rule of \secref{crossfit} accepts a proposal on fresh $A, B, P$ splits, and the iteration stops otherwise. 

\section{Experiments}
\seclabel{experiments}
 
We evaluate CoRP against RandOpt and gradient-based baselines on a shared benchmark and address four questions in turn. How much of the K=50 ensemble gain does a single CoRP model recover, and at what cost. When can a rewarded population actually be consolidated into one update. And what kinds of errors does the operator repair.
 
\textbf{Models.}
We use five instruction-tuned models that span 0.5B to 8B parameters and three pretraining lineages: Qwen2.5-0.5B, Qwen2.5-1.5B, and Qwen2.5-3B~\citep{qwen2024qwen25}, OLMo3-7B~\citep{olmo2025olmo3}, and Llama-3.1-8B~\citep{grattafiori2024llama3}.

\textbf{Tasks.}
Each model is evaluated on five tasks. Three of them test mathematical reasoning at increasing difficulty, Countdown~\citep{gandhi2024sos}, GSM8K~\citep{cobbe2021gsm8k}, and OlympiadBench~\citep{he2024olympiad}. ROCStories tests creative writing~\citep{mostafazadeh2016roc}, and MBPP tests short Python program synthesis~\citep{austin2021mbpp}.
 
\textbf{Perturbation population.}
For each model-task pair we sample $N{=}500$ rewarded perturbations using the noise-scale mixture $\Sigma = \{5\mathrm{e}{-}4, 1\mathrm{e}{-}3, 2\mathrm{e}{-}3\}$ from~\citet{gan2026neuralthickets}, and we follow the RandOpt evaluation protocol exactly. Each perturbation is stored by its random seed and noise scale and regenerated on demand, so the population's storage cost is independent of model size. The first 200 training examples are partitioned into the support folds $A$, $B$, and $P$ of \secref{crossfit}, and all results are reported on the original test set.

\begin{table*}[t]
\footnotesize
\caption{Comparison to post-training baselines. We report test-set accuracy, except for MBPP where we report pass@1.
\greedy{Blue} marks the overall best result when it is achieved by optimization-based methods.
\sample{Green} marks the best perturbation-based method.
}
\label{tab:CoRP_filtered_comparison}
\renewcommand{\arraystretch}{0.85}
\begin{center}
\begin{adjustbox}{width=0.9\textwidth}
\begin{tabular}{ll|ccccc}
\toprule
\multirow{2}{*}{Model} & \multirow{2}{*}{Method} 
& \textbf{Countdown} & \textbf{GSM8K} & \textbf{OlyBench} & \textbf{ROCStories} & \textbf{MBPP} \\
 &  & (math) & (math) & (math) & (writing) & (prog.) \\
\midrule
\rowcolor{paleviolet}\multirow{7}{*}{\cellcolor{white}\shortstack{Qwen2.5-0.5B-Inst}}
 & Base & 0.1$\pm$0.1 & 39.9$\pm$0.0 & 4.2$\pm$0.4 & 22.0$\pm$0.0 & 30.9$\pm$0.0 \\
 \cmidrule(l){2-7}
 & PPO & \greedy{14.8$\pm$2.8} & 43.2$\pm$1.2 & \greedy{16.1$\pm$0.5} & 19.1$\pm$0.0 & 37.8$\pm$3.0 \\
 & GRPO & 13.0$\pm$0.0 & 48.4$\pm$0.0 & 6.9$\pm$0.1 & 30.9$\pm$1.1 & 42.8$\pm$3.7 \\
 \cmidrule(l){2-7}
 & RandOpt ($K$=1)  & 4.8$\pm$0.8 & 44.8$\pm$0.5 & 8.6$\pm$0.7 & 23.3$\pm$0.9 & 30.7$\pm$0.7 \\
 & RandOpt ($K$=50) & 8.4$\pm$0.3 & \sample{54.1$\pm$0.8} & \sample{15.8$\pm$0.7} & \sample{32.2$\pm$0.4} & \sample{46.2$\pm$0.4} \\
 & CoRP & \sample{14.5$\pm$1.2} & 43.9$\pm$0.6 & 9.6$\pm$0.4 & 32.0$\pm$0.9 & 34.3$\pm$0.6 \\
\midrule
\rowcolor{paleviolet}\multirow{7}{*}{\cellcolor{white}\shortstack{Qwen2.5-1.5B-Inst}}
 & Base & 6.7$\pm$0.0 & 58.8$\pm$0.2 & 13.4$\pm$0.4 & 46.7$\pm$0.1 & 62.3$\pm$0.2 \\
 \cmidrule(l){2-7}
 & PPO & 27.0$\pm$0.0 & 71.6$\pm$0.7 & 26.3$\pm$0.1 & 51.8$\pm$0.8 & 69.5$\pm$0.4 \\
 & GRPO & 27.5$\pm$0.2 & 72.1$\pm$0.7 & 18.8$\pm$0.8 & \greedy{53.6$\pm$1.3} & \greedy{70.2$\pm$0.4} \\
 \cmidrule(l){2-7}
 & RandOpt ($K$=1)  & 9.1$\pm$0.3 & 59.1$\pm$0.8 & 19.9$\pm$0.7 & 46.6$\pm$0.2 & 61.1$\pm$0.6 \\
 & RandOpt ($K$=50) & \sample{52.7$\pm$0.6} & \sample{76.4$\pm$0.3} & \sample{30.4$\pm$0.7} & 48.5$\pm$0.7 & \sample{69.6$\pm$0.5} \\
 & CoRP & 26.2$\pm$0.7 & 70.1$\pm$0.3 & 20.5$\pm$0.5 & \sample{52.2$\pm$0.6} & 64.0$\pm$0.5 \\
\midrule
\rowcolor{paleviolet}\multirow{7}{*}{\cellcolor{white}\shortstack{Qwen2.5-3B-Inst}}
 & Base & 10.0$\pm$0.1 & 79.8$\pm$0.4 & 24.5$\pm$0.2 & 54.7$\pm$0.1 & 69.5$\pm$0.3 \\
 \cmidrule(l){2-7}
 & PPO & 35.3$\pm$0.1 & 83.1$\pm$0.2 & 34.4$\pm$0.2 & 49.0$\pm$0.6 & 76.3$\pm$1.0 \\
 & GRPO & 32.6$\pm$0.1 & 83.2$\pm$0.2 & 29.0$\pm$0.0 & 56.3$\pm$4.4 & \greedy{77.0$\pm$0.9} \\
 \cmidrule(l){2-7}
 & RandOpt ($K$=1)  & 10.4$\pm$0.8 & 81.4$\pm$0.9 & 30.3$\pm$0.4 & 54.6$\pm$0.4 & 69.1$\pm$1.1 \\
 & RandOpt ($K$=50) & \sample{58.4$\pm$0.2} & \sample{87.1$\pm$0.6} & \sample{39.2$\pm$0.6} & 56.5$\pm$0.3 & \sample{75.9$\pm$0.6} \\
 & CoRP & 36.9$\pm$0.3 & 82.3$\pm$0.4 & 32.5$\pm$0.5 & \sample{59.1$\pm$0.6} & 70.3$\pm$0.8 \\
\midrule
\rowcolor{paleviolet}\multirow{7}{*}{\cellcolor{white}\shortstack{OLMo3-7B-Inst}}
 & Base & 64.8$\pm$0.2 & 82.9$\pm$0.4 & 28.7$\pm$0.1 & 64.0$\pm$0.0 & 65.9$\pm$0.2 \\
 \cmidrule(l){2-7}
 & PPO & 69.0$\pm$0.0 & 88.4$\pm$0.4 & 28.0$\pm$0.2 & 64.7$\pm$0.6 & 67.7$\pm$0.6 \\
 & GRPO & 68.5$\pm$0.7 & 87.0$\pm$0.2 & 27.9$\pm$0.6 & \greedy{65.8$\pm$0.6} & 70.8$\pm$2.2 \\
 \cmidrule(l){2-7}
 & RandOpt ($K$=1)  & 74.4$\pm$0.7 & 83.2$\pm$0.6 & 37.5$\pm$0.3 & 64.2$\pm$0.8 & 66.5$\pm$0.6 \\
 & RandOpt ($K$=50) & \sample{85.0$\pm$0.2} & \sample{89.5$\pm$0.2} & 35.4$\pm$0.4 & 64.5$\pm$0.3 & \sample{75.1$\pm$0.9} \\
 & CoRP & 78.9$\pm$0.4 & 84.8$\pm$0.1 & \sample{39.2$\pm$0.6} & \sample{64.8$\pm$0.7} & 69.1$\pm$0.3 \\
\midrule
\rowcolor{paleviolet}\multirow{7}{*}{\cellcolor{white}\shortstack{Llama3.1-8B-Inst}}
 & Base & 10.8$\pm$0.2 & 79.8$\pm$0.3 & 19.2$\pm$0.3 & 51.8$\pm$0.3 & 56.4$\pm$0.0 \\
 \cmidrule(l){2-7}
 & PPO & 9.9$\pm$0.1 & 81.6$\pm$0.7 & 16.1$\pm$0.6 & 57.8$\pm$1.3 & 55.2$\pm$1.5 \\
 & GRPO & 10.0$\pm$0.2 & 80.2$\pm$1.1 & 23.7$\pm$0.9 & \greedy{62.2$\pm$2.8} & 61.0$\pm$1.9 \\
 \cmidrule(l){2-7}
 & RandOpt ($K$=1)  & 10.7$\pm$0.9 & 80.4$\pm$0.6 & 11.9$\pm$0.5 & 50.8$\pm$0.7 & 55.6$\pm$0.8 \\
 & RandOpt ($K$=50) & \sample{63.6$\pm$0.4} & \sample{86.7$\pm$0.6} & \sample{32.1$\pm$0.0} & 59.0$\pm$0.7 & \sample{65.2$\pm$0.9} \\
 & CoRP & 42.3$\pm$0.6 & 80.6$\pm$0.2 & 20.9$\pm$0.5 & \sample{60.5$\pm$0.4} & 61.1$\pm$0.8  \\
\bottomrule
\end{tabular}
\end{adjustbox}
\end{center}
\end{table*}

\subsection{Comparison to Standard Post-Training Methods}
\seclabel{main-comparison}

\Cref{tab:CoRP_filtered_comparison} compares CoRP with PPO, GRPO, and two RandOpt references. RandOpt $K{=}1$ keeps the best single perturbation, RandOpt $K{=}50$ keeps a prediction-level ensemble of the top fifty, and CoRP commits one consolidated update.

Compared with RandOpt $K{=}1$, CoRP improves 24 of 25 model-task pairs. It reaches or exceeds RandOpt $K{=}50$ in six pairs, including four ROCStories settings, Qwen2.5-0.5B on Countdown, and OLMo3-7B on OlympiadBench. PPO or GRPO gives the best result in a minority of cells. These results place CoRP between the best single perturbation and the full ensemble. It usually extracts more deployable signal than RandOpt $K{=}1$, while leaving a gap to $K{=}50$ in settings where specialists remain hard to compress.

\begin{wraptable}{r}{0.38\columnwidth}
\footnotesize
\vspace{-0.6cm}
\renewcommand{\arraystretch}{0.8}
\caption{Post-training cost in model-example forward equivalents, inference passes, and average score over 25 model-task pairs. RandOpt uses $N{=}5000$ and CoRP uses $N{=}500$. CoRP cost includes all stages.}
\label{tab:compute}
\begin{center}
\begin{adjustbox}{width=\linewidth,center}
\begin{tabular}{@{}lccc@{}}
\toprule
Method & Post-train & Inf. & Score \\
\midrule
Base             & --                         & $1$           & $41.9$ \\
\midrule
PPO              & $1{\times}10^6$           & $1$           & $47.7$ \\
GRPO             & $1{\times}10^6$           & $1$           & $48.8$ \\
\midrule
RandOpt $K{=}1$  & $1{\times}10^6$           & $1$           & $43.5$ \\
RandOpt $K{=}50$ & $1{\times}10^6$           & $50$          & $\mathbf{56.3}$ \\
\midrule
\textbf{CoRP}   & $\mathbf{1.46{\times}10^5}$ & $\mathbf{1}$ & $\underline{50.0}$ \\
\bottomrule
\end{tabular}
\end{adjustbox}
\end{center}
\vspace{-0.4em}
\end{wraptable}

\Cref{tab:compute} reports the cost at which each method reaches its average score. We measure cost in model-example forward equivalents, omitting the common $2PL$ factor. RandOpt scores $N{=}5000$ perturbations on $200$ support examples, for a total of $1{\times}10^6$ forward equivalents. CoRP follows the same sampling procedure with $N{=}500$, while candidate evaluation, validation, probe calibration, and local search bring its complete cost from $1{\times}10^5$ to $1.46{\times}10^5$ forward equivalents, or $14.6\%$ of the RandOpt budget. PPO and GRPO remain matched to RandOpt $K{=}50$ under the same protocol. CoRP improves the base model by $8.1$ points, exceeds single-inference RandOpt by $6.5$ points, and recovers more than half of the $K{=}50$ ensemble gain. RandOpt $K{=}50$ retains the highest score through fifty inference passes, while CoRP recovers much of that gain with one deployed model and one pass per example.

\subsection{When Can a Rewarded Population Be Consolidated?}
\seclabel{when-consolidates}

A prediction-level ensemble can keep rewarded perturbations separate until inference. A consolidated model cannot. It has to write their useful parts into one set of weights, and any incompatible structure that comes along will be written in too.

We make this tension visible by sweeping two perturbations at a time. For rewarded perturbations $\Delta_i$ and $\Delta_j$, we evaluate $\theta_0 + \eta(\alpha \Delta_i + \beta \Delta_j)$ over a grid of mixing coefficients $(\alpha,\beta)$. The resulting accuracy landscape shows which mixtures survive consolidation.

\begin{figure}[htbp]
    \centering
    \vspace{-0.2cm}
    \includegraphics[width=\textwidth]{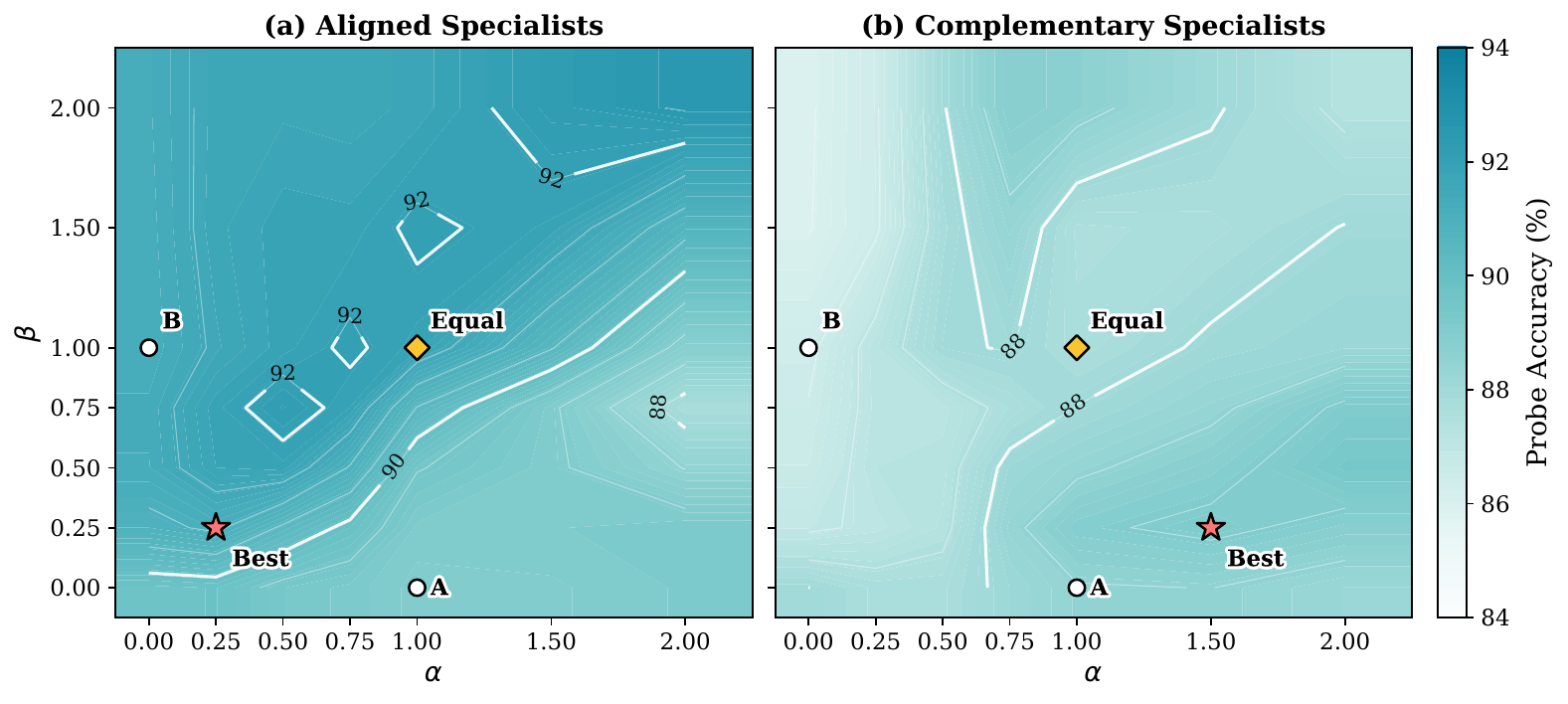}
    \vspace{-0.4cm}
    \caption{Pairwise consolidation landscapes on Qwen2.5-3B-Instruct and GSM8K. The contours visualize probe accuracy for $\theta_0 + \eta(\alpha \Delta_i + \beta \Delta_j)$. Markers A and B are the individual perturbations, Equal is equal weighting, Best is the best probe point. Panel (a) shows an aligned pair with a broad high-accuracy region. Panel (b) shows a complementary pair with a narrower region and a stronger reweighted point.}
    \vspace{-0.2cm}
    \label{fig:pairwise_landscape}
\end{figure}

\Cref{fig:pairwise_landscape} shows two characteristic cases on Qwen2.5-3B-Instruct and GSM8K. In panel (a), the two perturbations repair the same probe errors. The high-accuracy region is broad, and equal weighting already lands inside it. This pair behaves like two samples of the same consolidatable component. In panel (b) the repair pattern changes. The two perturbations fix mostly different examples. The high-accuracy region shrinks, and equal weighting falls below the best reweighted point. Complementary perturbations can still be consolidated, but only under more specific weights, and the room for error is smaller. This is the target of CoRP, to preserve the compatible part of the rewarded population, not every specialist that an ensemble can use separately.

\subsection{Composition of the Gain}
\seclabel{gain-composition}

The same score can hide different behavior. A method may solve examples the base model missed, or it may make an already-correct solution easier for the strict evaluator to parse. It may also improve some examples while regressing on others. Following~\citet{gan2026neuralthickets}, we decompose the test-set outcome into four categories. A \emph{retained correct} example is strictly correct for the base model and remains correct after adaptation. A \emph{format fix} is one where the base produces the right answer content but fails the strict format and the adapted model becomes strictly correct. A \emph{reasoning fix} is one where the base fails on the answer content and the adapted model becomes strictly correct. A \emph{regression} is strictly correct for the base and incorrect after adaptation. Precise definitions and the format-vs-answer distinction per task are in the appendix.

\begin{figure}[htbp]
    \centering
    \vspace{-0.2cm}
    \includegraphics[width=1\textwidth]{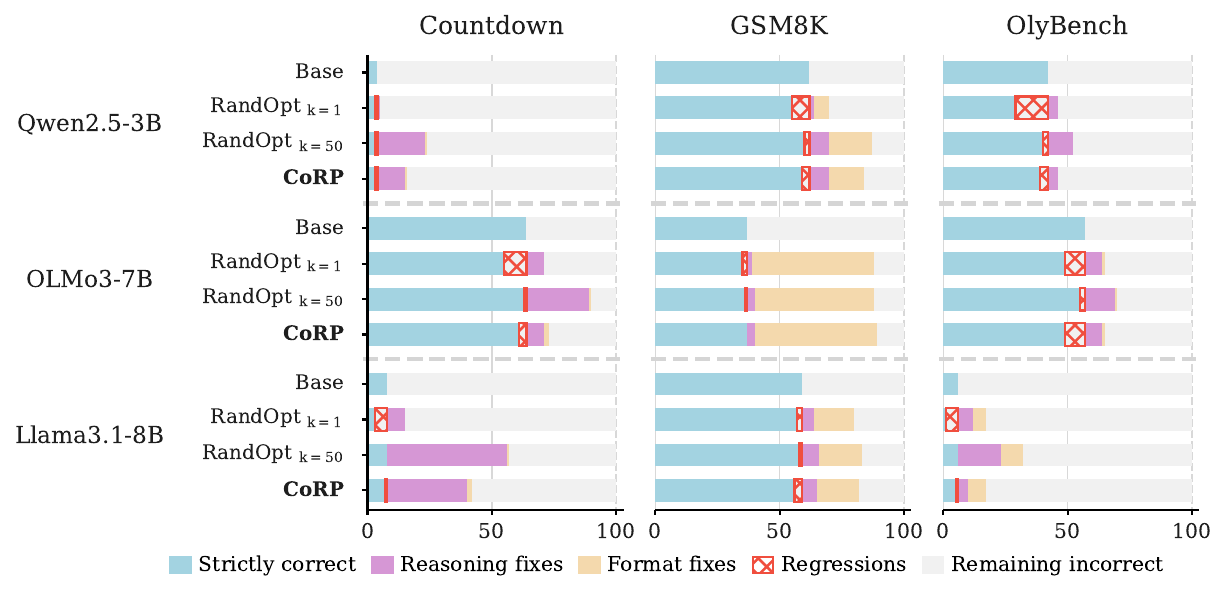}
    \vspace{-0.6cm}
    \caption{Composition of the test-set outcome on the three math tasks. Each row is one method, and each segment is the fraction of test examples that fall into one of four categories defined in the text. Bars are stacked left to right by strictly correct examples, reasoning fixes, and format fixes. Regressions are overlaid as a hatched negative segment.}
    \label{fig:composition}
\end{figure}

\Cref{fig:composition} shows that CoRP's gain is not only format repair. GSM8K does carry a visible format component for the stronger base models, but Countdown and OlympiadBench show clear reasoning fixes, where the base fails on the answer content and CoRP makes the example strictly correct. This matters because CoRP performs one inference with one updated model, rather than relying on a vote across perturbed models.

The same panels also show why consolidating several rewarded perturbations is different from selecting one. RandOpt $K{=}1$ introduces visible regressions, especially on Countdown and OlympiadBench. CoRP keeps the single-model deployment of $K{=}1$, but its regression segments are smaller in most panels. RandOpt $K{=}50$ can vote regressions away at inference. CoRP cannot. Reducing them before deployment is a central part of consolidation, and the held-out gate of \secref{crossfit} is what does it.

\paragraph{Robustness checks.}
Each component of CoRP carries weight, and full CoRP improves on naive baselines including reward-weighted averaging and direct top-$r$ subspace projection. The choice of compatibility weights $\gamma_a, \gamma_d$ is also not fragile in a sweep over $\{0.1, 0.5, 1, 2, 5\}^2$ on Qwen2.5-3B / GSM8K. We report ablations, naive baselines, and the sensitivity sweep in Appendix~\secref{add_results}.

\section{Related Work}
\seclabel{related}
 
\textbf{Search in weight space.}
CoRP is closest in spirit to a line of self-elicitation post-training methods that sample from the model and consolidate the exposure back into the weights, including RLVR and on-policy distillation. Within this lineage, a body of work treats LLM post-training as a search problem in weight space rather than as gradient descent on a loss. Evolution strategies and the cross-entropy method use reward-weighted samples to update a sampling distribution over weights~\citep{salimans2017es, mannor2003cem, rubinstein1999cem, peters2007rwr, peng2019awr}, with covariance-adaptive variants concentrating exploration in informative directions~\citep{hansen2016cmaes, maheswaranathan2018guidedes, choromanski2019asebo}. RandOpt~\citep{gan2026neuralthickets} pushes this idea to its extreme by removing iterative search entirely. A single round of rewarded Gaussian perturbations with prediction-level ensembling already matches PPO and GRPO on contemporary models. CoRP shares RandOpt's reliance on a single rewarded sampling stage as the primary source of perturbations, but consolidates the rewarded population into one deployable update rather than carrying it forward as an ensemble.
 
\textbf{Parameter-space model merging.}
A second relevant line aggregates several models in parameter space. Model soups average independently fine-tuned checkpoints that share a low-loss basin~\citep{wortsman2022soups, frankle2020linearmode}. When the sources disagree, geometry-aware corrections become necessary. TIES-merging trims small magnitudes and resolves sign conflicts~\citep{yadav2023ties}, DARE drops and rescales parameters to cut interference~\citep{yu2024dare}, Fisher-weighted merging weights sources by local curvature~\citep{matena2022fisher}, and task arithmetic operates on task vectors anchored to a shared base~\citep{ilharco2023task}. CoRP applies the same principle to a different source family. Its sources are random rewarded perturbations around one pretrained model rather than independently fine-tuned checkpoints, so compatibility is measured by alignment and dispersion with respect to a provisional reward-weighted mean rather than by sign, magnitude, or curvature.
 
\textbf{Low-dimensional structure in LLM adaptation.}
The stable low-rank signal we measure in rewarded perturbations aligns with a larger body of work on low-dimensional LLM adaptation. Intrinsic-dimension analyses show that fine-tuning succeeds within a small random subspace~\citep{aghajanyan2020intrinsicdim}. LoRA~\citep{hu2022lora} and its variants restrict updates to low-rank components without sacrificing accuracy. \citet{liang2026blessing} attribute the success of LLM adaptation to a low-dimensional curvature geometry, and~\citet{morris2026thirteen} show that a mathematical reasoning task can be learned by updating thirteen parameters. Our characterization adds a complementary observation at a different granularity, namely that reward-driven random search around a pretrained model concentrates its useful variance on a consistent low-rank subspace of a coordinate sketch space.

\section{Discussion}
\seclabel{discussion}
 
Post-training can be viewed as an elicitation loop. Output-space methods sample trajectories and use reward or teacher feedback to write useful behavior back into the model~\citep{shao2024grpo, agarwal2024onpolicykd}. Evolution strategies offer a complementary route by using rewards from weight perturbations to update model parameters~\citep{qiu2026evolutionstrategies}. RandOpt reveals a related opportunity in weight space, but retains rewarded specialists for inference-time ensembling~\citep{gan2026neuralthickets}. CoRP shows that useful variation across these specialists can instead be consolidated into a single model.

Across all $25$ model-task pairs, rewarded perturbations exhibit reproducible low-rank structure even though a stable mean direction is absent in most cases. This suggests that shared geometry need not imply a common update direction. CoRP uses compatibility-aware consolidation to combine rewarded perturbations into a deployable update without any gradient through the language model. It recovers more than half of the $50$-pass ensemble gain at $14.6\%$ of RandOpt's post-training cost and one forward pass at inference. More broadly, the result connects weight-space self-elicitation to evidence that language model adaptation often admits low-dimensional structure~\citep{aghajanyan2020intrinsicdim, liang2026blessing}. The rewarded neighborhood of a well-pretrained model is therefore not only dense with specialists but also contains enough shared geometry to improve a single model.

\paragraph{Limitations \& Future Work.}
CoRP currently targets settings where a reliable scalar reward is available on a small support set. Our experiments establish consolidation on structured or verifiable tasks. Whether the same population structure can be recovered from preference-based feedback, noisier rewards, or long-horizon outcomes remains open. These settings preserve the underlying consolidation problem, but make it harder to identify which perturbations carry reliable signal.

A second boundary is complementarity. When rewarded perturbations repair different examples, an inference-time ensemble can retain them separately, whereas one checkpoint must reconcile them within a shared parameter region. CoRP captures the compatible component of such populations, but does not preserve every specialist that the ensemble can use. This gap motivates consolidation operators that first represent multiple behavioral modes and then determine which of them can coexist in one update. The validation gate could similarly be extended beyond the target reward to protect unrelated capabilities during consolidation.

Together, these results establish rewarded perturbation populations as a meaningful object for post-training. Their geometry reveals which parts of local weight-space variation are shared, while consolidation tests whether that shared structure can survive in a single model. CoRP provides a first step toward turning this population-level structure into a practical model update.



\clearpage
{
\bibliographystyle{unsrtnat}
\bibliography{references}

@misc{schulman2017ppo,
      title={Proximal Policy Optimization Algorithms}, 
      author={John Schulman and Filip Wolski and Prafulla Dhariwal and Alec Radford and Oleg Klimov},
      year={2017},
      eprint={1707.06347},
      archivePrefix={arXiv},
      primaryClass={cs.LG},
      url={https://arxiv.org/abs/1707.06347}, 
}

@inproceedings{ouyang2022rlhf,
 author = {Ouyang, Long and Wu, Jeffrey and Jiang, Xu and Almeida, Diogo and Wainwright, Carroll and Mishkin, Pamela and Zhang, Chong and Agarwal, Sandhini and Slama, Katarina and Ray, Alex and Schulman, John and Hilton, Jacob and Kelton, Fraser and Miller, Luke and Simens, Maddie and Askell, Amanda and Welinder, Peter and Christiano, Paul F and Leike, Jan and Lowe, Ryan},
 booktitle = {Advances in Neural Information Processing Systems},
 editor = {S. Koyejo and S. Mohamed and A. Agarwal and D. Belgrave and K. Cho and A. Oh},
 pages = {27730--27744},
 publisher = {Curran Associates, Inc.},
 title = {Training language models to follow instructions with human feedback},
 url = {https://proceedings.neurips.cc/paper_files/paper/2022/file/b1efde53be364a73914f58805a001731-Paper-Conference.pdf},
 volume = {35},
 year = {2022}
}

@article{shao2024grpo,
  title={Deepseekmath: Pushing the limits of mathematical reasoning in open language models},
  author={Shao, Zhihong and Wang, Peiyi and Zhu, Qihao and Xu, Runxin and Song, Junxiao and Bi, Xiao and Zhang, Haowei and Zhang, Mingchuan and Li, YK and Wu, Yang and others},
  journal={arXiv preprint arXiv:2402.03300},
  year={2024}
}

@article{deepseek2025r1,
  title={DeepSeek-R1 incentivizes reasoning in LLMs through reinforcement learning},
  author={Guo, Daya and Yang, Dejian and Zhang, Haowei and Song, Junxiao and Wang, Peiyi and Zhu, Qihao and Xu, Runxin and Zhang, Ruoyu and Ma, Shirong and Bi, Xiao and others},
  journal={Nature},
  volume={645},
  number={8081},
  pages={633--638},
  year={2025},
  publisher={Nature Publishing Group UK London}
}

@misc{shao2025spurious,
      title={Spurious Rewards: Rethinking Training Signals in RLVR}, 
      author={Rulin Shao and Shuyue Stella Li and Rui Xin and Scott Geng and Yiping Wang and Sewoong Oh and Simon Shaolei Du and Nathan Lambert and Sewon Min and Ranjay Krishna and Yulia Tsvetkov and Hannaneh Hajishirzi and Pang Wei Koh and Luke Zettlemoyer},
      year={2026},
      eprint={2506.10947},
      archivePrefix={arXiv},
      primaryClass={cs.AI},
      url={https://arxiv.org/abs/2506.10947}, 
}

@misc{ward2025reasoning,
      title={Reasoning-Finetuning Repurposes Latent Representations in Base Models}, 
      author={Jake Ward and Chuqiao Lin and Constantin Venhoff and Neel Nanda},
      year={2025},
      eprint={2507.12638},
      archivePrefix={arXiv},
      primaryClass={cs.LG},
      url={https://arxiv.org/abs/2507.12638}, 
}

@inproceedings{
wei2022cot,
title={Chain of Thought Prompting Elicits Reasoning in Large Language Models},
author={Jason Wei and Xuezhi Wang and Dale Schuurmans and Maarten Bosma and brian ichter and Fei Xia and Ed H. Chi and Quoc V Le and Denny Zhou},
booktitle={Advances in Neural Information Processing Systems},
editor={Alice H. Oh and Alekh Agarwal and Danielle Belgrave and Kyunghyun Cho},
year={2022},
url={https://openreview.net/forum?id=_VjQlMeSB_J}
}

@inproceedings{
kojima2022zeroshot,
title={Large Language Models are Zero-Shot Reasoners},
author={Takeshi Kojima and Shixiang Shane Gu and Machel Reid and Yutaka Matsuo and Yusuke Iwasawa},
booktitle={Advances in Neural Information Processing Systems},
editor={Alice H. Oh and Alekh Agarwal and Danielle Belgrave and Kyunghyun Cho},
year={2022},
url={https://openreview.net/forum?id=e2TBb5y0yFf}
}

@inproceedings{
agarwal2024onpolicykd,
title={On-Policy Distillation of Language Models: Learning from Self-Generated Mistakes},
author={Rishabh Agarwal and Nino Vieillard and Yongchao Zhou and Piotr Stanczyk and Sabela Ramos Garea and Matthieu Geist and Olivier Bachem},
booktitle={The Twelfth International Conference on Learning Representations},
year={2024},
url={https://openreview.net/forum?id=3zKtaqxLhW}
}

@misc{yang2025qwen3,
      title={Qwen3 Technical Report}, 
      author={An Yang and Anfeng Li and Baosong Yang and Beichen Zhang and Binyuan Hui and Bo Zheng and Bowen Yu and Chang Gao and Chengen Huang and Chenxu Lv and Chujie Zheng and Dayiheng Liu and Fan Zhou and Fei Huang and Feng Hu and Hao Ge and Haoran Wei and Huan Lin and Jialong Tang and Jian Yang and Jianhong Tu and Jianwei Zhang and Jianxin Yang and Jiaxi Yang and Jing Zhou and Jingren Zhou and Junyang Lin and Kai Dang and Keqin Bao and Kexin Yang and Le Yu and Lianghao Deng and Mei Li and Mingfeng Xue and Mingze Li and Pei Zhang and Peng Wang and Qin Zhu and Rui Men and Ruize Gao and Shixuan Liu and Shuang Luo and Tianhao Li and Tianyi Tang and Wenbiao Yin and Xingzhang Ren and Xinyu Wang and Xinyu Zhang and Xuancheng Ren and Yang Fan and Yang Su and Yichang Zhang and Yinger Zhang and Yu Wan and Yuqiong Liu and Zekun Wang and Zeyu Cui and Zhenru Zhang and Zhipeng Zhou and Zihan Qiu},
      year={2025},
      eprint={2505.09388},
      archivePrefix={arXiv},
      primaryClass={cs.CL},
      url={https://arxiv.org/abs/2505.09388}, 
}

@misc{xiao2026mimo,
      title={MiMo-V2-Flash Technical Report}, 
      author={Core Team and Bangjun Xiao and Bingquan Xia and Bo Yang and Bofei Gao and Bowen Shen and Chen Zhang and Chenhong He and Chiheng Lou and Fuli Luo and Gang Wang and Gang Xie and Hailin Zhang and Hanglong Lv and Hanyu Li and Heyu Chen and Hongshen Xu and Houbin Zhang and Huaqiu Liu and Jiangshan Duo and Jianyu Wei and Jiebao Xiao and Jinhao Dong and Jun Shi and Junhao Hu and Kainan Bao and Kang Zhou and Lei Li and Liang Zhao and Linghao Zhang and Peidian Li and Qianli Chen and Shaohui Liu and Shihua Yu and Shijie Cao and Shimao Chen and Shouqiu Yu and Shuo Liu and Tianling Zhou and Weijiang Su and Weikun Wang and Wenhan Ma and Xiangwei Deng and Bohan Mao and Bowen Ye and Can Cai and Chenghua Wang and Chengxuan Zhu and Chong Ma and Chun Chen and Chunan Li and Dawei Zhu and Deshan Xiao and Dong Zhang and Duo Zhang and Fangyue Liu and Feiyu Yang and Fengyuan Shi and Guoan Wang and Hao Tian and Hao Wu and Heng Qu and Hongfei Yi and Hongxu An and Hongyi Guan and Xing Zhang and Yifan Song and Yihan Yan and Yihao Zhao and Yingchun Lai and Yizhao Gao and Yu Cheng and Yuanyuan Tian and Yudong Wang and Zhen Tang and Zhengju Tang and Zhengtao Wen and Zhichao Song and Zhixian Zheng and Zihan Jiang and Jian Wen and Jiarui Sun and Jiawei Li and Jinlong Xue and Jun Xia and Kai Fang and Menghang Zhu and Nuo Chen and Qian Tu and Qihao Zhang and Qiying Wang and Rang Li and Rui Ma and Shaolei Zhang and Shengfan Wang and Shicheng Li and Shuhao Gu and Shuhuai Ren and Sirui Deng and Tao Guo and Tianyang Lu and Weiji Zhuang and Weikang Zhang and Weimin Xiong and Wenshan Huang and Wenyu Yang and Xin Zhang and Xing Yong and Xu Wang and Xueyang Xie and Yilin Jiang and Yixin Yang and Yongzhe He and Yu Tu and Yuanliang Dong and Yuchen Liu and Yue Ma and Yue Yu and Yuxing Xiang and Zhaojun Huang and Zhenru Lin and Zhipeng Xu and Zhiyang Chen and Zhonghua Deng and Zihan Zhang and Zihao Yue},
      year={2026},
      eprint={2601.02780},
      archivePrefix={arXiv},
      primaryClass={cs.CL},
      url={https://arxiv.org/abs/2601.02780}, 
}

@misc{gan2026neuralthickets,
      title={Neural Thickets: Diverse Task Experts Are Dense Around Pretrained Weights}, 
      author={Yulu Gan and Phillip Isola},
      year={2026},
      eprint={2603.12228},
      archivePrefix={arXiv},
      primaryClass={cs.LG},
      url={https://arxiv.org/abs/2603.12228}, 
}

@misc{salimans2017es,
      title={Evolution Strategies as a Scalable Alternative to Reinforcement Learning}, 
      author={Tim Salimans and Jonathan Ho and Xi Chen and Szymon Sidor and Ilya Sutskever},
      year={2017},
      eprint={1703.03864},
      archivePrefix={arXiv},
      primaryClass={stat.ML},
      url={https://arxiv.org/abs/1703.03864}, 
}

@misc{hansen2016cmaes,
      title={The CMA Evolution Strategy: A Tutorial}, 
      author={Nikolaus Hansen},
      year={2023},
      eprint={1604.00772},
      archivePrefix={arXiv},
      primaryClass={cs.LG},
      url={https://arxiv.org/abs/1604.00772}, 
}

@InProceedings{maheswaranathan2018guidedes,
  title = 	 {Guided evolutionary strategies: augmenting random search with surrogate gradients},
  author =       {Maheswaranathan, Niru and Metz, Luke and Tucker, George and Choi, Dami and Sohl-Dickstein, Jascha},
  booktitle = 	 {Proceedings of the 36th International Conference on Machine Learning},
  pages = 	 {4264--4273},
  year = 	 {2019},
  editor = 	 {Chaudhuri, Kamalika and Salakhutdinov, Ruslan},
  volume = 	 {97},
  series = 	 {Proceedings of Machine Learning Research},
  month = 	 {09--15 Jun},
  publisher =    {PMLR},
  url = 	 {https://proceedings.mlr.press/v97/maheswaranathan19a.html}
}

@inproceedings{choromanski2019asebo,
 author = {Choromanski, Krzysztof M and Pacchiano, Aldo and Parker-Holder, Jack and Tang, Yunhao and Sindhwani, Vikas},
 booktitle = {Advances in Neural Information Processing Systems},
 editor = {H. Wallach and H. Larochelle and A. Beygelzimer and F. d\textquotesingle Alch\'{e}-Buc and E. Fox and R. Garnett},
 pages = {},
 publisher = {Curran Associates, Inc.},
 title = {From Complexity to Simplicity: Adaptive ES-Active Subspaces for Blackbox Optimization},
 url = {https://proceedings.neurips.cc/paper_files/paper/2019/file/88bade49e98db8790df275fcebb37a13-Paper.pdf},
 volume = {32},
 year = {2019}
}

@inproceedings{peters2007rwr,
author = {Peters, Jan and Schaal, Stefan},
title = {Reinforcement learning by reward-weighted regression for operational space control},
year = {2007},
isbn = {9781595937933},
publisher = {Association for Computing Machinery},
address = {New York, NY, USA},
url = {https://doi.org/10.1145/1273496.1273590},
doi = {10.1145/1273496.1273590},
booktitle = {Proceedings of the 24th International Conference on Machine Learning},
pages = {745–750},
numpages = {6},
location = {Corvalis, Oregon, USA},
series = {ICML '07}
}

@misc{peng2019awr,
      title={Advantage-Weighted Regression: Simple and Scalable Off-Policy Reinforcement Learning}, 
      author={Xue Bin Peng and Aviral Kumar and Grace Zhang and Sergey Levine},
      year={2019},
      eprint={1910.00177},
      archivePrefix={arXiv},
      primaryClass={cs.LG},
      url={https://arxiv.org/abs/1910.00177}, 
}

@article{rubinstein1999cem,
  title={The cross-entropy method for combinatorial and continuous optimization},
  author={Rubinstein, Reuven},
  journal={Methodology and computing in applied probability},
  volume={1},
  number={2},
  pages={127--190},
  year={1999},
  publisher={Springer}
}

@inproceedings{mannor2003cem,
author = {Mannor, Shie and Rubinstein, Reuven and Gat, Yohai},
title = {The cross entropy method for fast policy search},
year = {2003},
isbn = {1577351894},
publisher = {AAAI Press},
booktitle = {Proceedings of the Twentieth International Conference on International Conference on Machine Learning},
pages = {512–519},
numpages = {8},
location = {Washington, DC, USA},
series = {ICML'03}
}

@InProceedings{wortsman2022soups,
  title = 	 {Model soups: averaging weights of multiple fine-tuned models improves accuracy without increasing inference time},
  author =       {Wortsman, Mitchell and Ilharco, Gabriel and Gadre, Samir Ya and Roelofs, Rebecca and Gontijo-Lopes, Raphael and Morcos, Ari S and Namkoong, Hongseok and Farhadi, Ali and Carmon, Yair and Kornblith, Simon and Schmidt, Ludwig},
  booktitle = 	 {Proceedings of the 39th International Conference on Machine Learning},
  pages = 	 {23965--23998},
  year = 	 {2022},
  editor = 	 {Chaudhuri, Kamalika and Jegelka, Stefanie and Song, Le and Szepesvari, Csaba and Niu, Gang and Sabato, Sivan},
  volume = 	 {162},
  series = 	 {Proceedings of Machine Learning Research},
  month = 	 {17--23 Jul},
  publisher =    {PMLR},
  url = 	 {https://proceedings.mlr.press/v162/wortsman22a.html}
}

@InProceedings{frankle2020linearmode,
  title = 	 {Linear Mode Connectivity and the Lottery Ticket Hypothesis},
  author =       {Frankle, Jonathan and Dziugaite, Gintare Karolina and Roy, Daniel and Carbin, Michael},
  booktitle = 	 {Proceedings of the 37th International Conference on Machine Learning},
  pages = 	 {3259--3269},
  year = 	 {2020},
  editor = 	 {III, Hal Daumé and Singh, Aarti},
  volume = 	 {119},
  series = 	 {Proceedings of Machine Learning Research},
  month = 	 {13--18 Jul},
  publisher =    {PMLR},
  url = 	 {https://proceedings.mlr.press/v119/frankle20a.html}
}

@inproceedings{
ilharco2023task,
title={Editing models with task arithmetic},
author={Gabriel Ilharco and Marco Tulio Ribeiro and Mitchell Wortsman and Ludwig Schmidt and Hannaneh Hajishirzi and Ali Farhadi},
booktitle={The Eleventh International Conference on Learning Representations },
year={2023},
url={https://openreview.net/forum?id=6t0Kwf8-jrj}
}

@inproceedings{
matena2022fisher,
title={Merging Models with Fisher-Weighted Averaging},
author={Michael S Matena and Colin Raffel},
booktitle={Advances in Neural Information Processing Systems},
editor={Alice H. Oh and Alekh Agarwal and Danielle Belgrave and Kyunghyun Cho},
year={2022},
url={https://openreview.net/forum?id=LSKlp_aceOC}
}

@inproceedings{
yadav2023ties,
title={{TIES}-Merging: Resolving Interference When Merging Models},
author={Prateek Yadav and Derek Tam and Leshem Choshen and Colin Raffel and Mohit Bansal},
booktitle={Thirty-seventh Conference on Neural Information Processing Systems},
year={2023},
url={https://openreview.net/forum?id=xtaX3WyCj1}
}

@inproceedings{
yu2024dare,
title={Language Models are Super Mario: Absorbing Abilities from Homologous Models as a Free Lunch},
author={Le Yu and Bowen Yu and Haiyang Yu and Fei Huang and Yongbin Li},
booktitle={Forty-first International Conference on Machine Learning},
year={2024},
url={https://openreview.net/forum?id=fq0NaiU8Ex}
}

@inproceedings{aghajanyan2020intrinsicdim,
    title = "Intrinsic Dimensionality Explains the Effectiveness of Language Model Fine-Tuning",
    author = "Aghajanyan, Armen  and
      Gupta, Sonal  and
      Zettlemoyer, Luke",
    editor = "Zong, Chengqing  and
      Xia, Fei  and
      Li, Wenjie  and
      Navigli, Roberto",
    booktitle = "Proceedings of the 59th Annual Meeting of the Association for Computational Linguistics and the 11th International Joint Conference on Natural Language Processing (Volume 1: Long Papers)",
    month = aug,
    year = "2021",
    address = "Online",
    publisher = "Association for Computational Linguistics",
    url = "https://aclanthology.org/2021.acl-long.568/",
    doi = "10.18653/v1/2021.acl-long.568",
    pages = "7319--7328"
}

@inproceedings{
hu2022lora,
title={Lo{RA}: Low-Rank Adaptation of Large Language Models},
author={Edward J Hu and yelong shen and Phillip Wallis and Zeyuan Allen-Zhu and Yuanzhi Li and Shean Wang and Lu Wang and Weizhu Chen},
booktitle={International Conference on Learning Representations},
year={2022},
url={https://openreview.net/forum?id=nZeVKeeFYf9}
}

@misc{liang2026blessing,
      title={The Blessing of Dimensionality in LLM Fine-tuning: A Variance-Curvature Perspective}, 
      author={Qiyao Liang and Jinyeop Song and Yizhou Liu and Jeff Gore and Ila Fiete and Risto Miikkulainen and Xin Qiu},
      year={2026},
      eprint={2602.00170},
      archivePrefix={arXiv},
      primaryClass={cs.LG},
      url={https://arxiv.org/abs/2602.00170}, 
}

@misc{morris2026thirteen,
      title={Learning to Reason in 13 Parameters}, 
      author={John X. Morris and Niloofar Mireshghallah and Mark Ibrahim and Saeed Mahloujifar},
      year={2026},
      eprint={2602.04118},
      archivePrefix={arXiv},
      primaryClass={cs.LG},
      url={https://arxiv.org/abs/2602.04118}, 
}

@article{stone1974cv,
  title={Cross-validatory choice and assessment of statistical predictions},
  author={Stone, Mervyn},
  journal={Journal of the royal statistical society: Series B (Methodological)},
  volume={36},
  number={2},
  pages={111--133},
  year={1974},
  publisher={Wiley Online Library}
}

@inproceedings{
chernozhukov2018doubleml,
title={Double/Debiased Machine Learning for Dynamic Treatment Effects},
author={Greg Lewis and Vasilis Syrgkanis},
booktitle={Advances in Neural Information Processing Systems},
editor={A. Beygelzimer and Y. Dauphin and P. Liang and J. Wortman Vaughan},
year={2021},
url={https://openreview.net/forum?id=StKuQ0-dltN}
}

@article{brown1910split,
  title={Some experimental results in the correlation of mental abilities 1},
  author={Brown, William},
  journal={British Journal of Psychology, 1904-1920},
  volume={3},
  number={3},
  pages={296--322},
  year={1910},
  publisher={Blackwell Publishing Ltd Oxford, UK}
}

@article{spearman1910split,
  title={Correlation calculated from faulty data},
  author={Spearman, Charles},
  journal={British journal of psychology},
  volume={3},
  number={3},
  pages={271},
  year={1910},
  publisher={Cambridge University Press}
}

@book{vershynin2018hdp,
  title={High-dimensional probability: An introduction with applications in data science},
  author={Vershynin, Roman},
  volume={47},
  year={2018},
  publisher={Cambridge university press}
}

@misc{qwen2024qwen25,
      title={Qwen2.5 Technical Report}, 
      author={Qwen and : and An Yang and Baosong Yang and Beichen Zhang and Binyuan Hui and Bo Zheng and Bowen Yu and Chengyuan Li and Dayiheng Liu and Fei Huang and Haoran Wei and Huan Lin and Jian Yang and Jianhong Tu and Jianwei Zhang and Jianxin Yang and Jiaxi Yang and Jingren Zhou and Junyang Lin and Kai Dang and Keming Lu and Keqin Bao and Kexin Yang and Le Yu and Mei Li and Mingfeng Xue and Pei Zhang and Qin Zhu and Rui Men and Runji Lin and Tianhao Li and Tianyi Tang and Tingyu Xia and Xingzhang Ren and Xuancheng Ren and Yang Fan and Yang Su and Yichang Zhang and Yu Wan and Yuqiong Liu and Zeyu Cui and Zhenru Zhang and Zihan Qiu},
      year={2025},
      eprint={2412.15115},
      archivePrefix={arXiv},
      primaryClass={cs.CL},
      url={https://arxiv.org/abs/2412.15115}, 
}

@misc{olmo2025olmo3,
      title={Olmo 3}, 
      author={Team Olmo and : and Allyson Ettinger and Amanda Bertsch and Bailey Kuehl and David Graham and David Heineman and Dirk Groeneveld and Faeze Brahman and Finbarr Timbers and Hamish Ivison and Jacob Morrison and Jake Poznanski and Kyle Lo and Luca Soldaini and Matt Jordan and Mayee Chen and Michael Noukhovitch and Nathan Lambert and Pete Walsh and Pradeep Dasigi and Robert Berry and Saumya Malik and Saurabh Shah and Scott Geng and Shane Arora and Shashank Gupta and Taira Anderson and Teng Xiao and Tyler Murray and Tyler Romero and Victoria Graf and Akari Asai and Akshita Bhagia and Alexander Wettig and Alisa Liu and Aman Rangapur and Chloe Anastasiades and Costa Huang and Dustin Schwenk and Harsh Trivedi and Ian Magnusson and Jaron Lochner and Jiacheng Liu and Lester James V. Miranda and Maarten Sap and Malia Morgan and Michael Schmitz and Michal Guerquin and Michael Wilson and Regan Huff and Ronan Le Bras and Rui Xin and Rulin Shao and Sam Skjonsberg and Shannon Zejiang Shen and Shuyue Stella Li and Tucker Wilde and Valentina Pyatkin and Will Merrill and Yapei Chang and Yuling Gu and Zhiyuan Zeng and Ashish Sabharwal and Luke Zettlemoyer and Pang Wei Koh and Ali Farhadi and Noah A. Smith and Hannaneh Hajishirzi},
      year={2026},
      eprint={2512.13961},
      archivePrefix={arXiv},
      primaryClass={cs.CL},
      url={https://arxiv.org/abs/2512.13961}, 
}

@misc{grattafiori2024llama3,
      title={The Llama 3 Herd of Models}, 
      author={Aaron Grattafiori and Abhimanyu Dubey and Abhinav Jauhri and Abhinav Pandey and Abhishek Kadian and Ahmad Al-Dahle and Aiesha Letman and Akhil Mathur and Alan Schelten and Alex Vaughan and Amy Yang and Angela Fan and Anirudh Goyal and Anthony Hartshorn and Aobo Yang and Archi Mitra and Archie Sravankumar and Artem Korenev and Arthur Hinsvark and Arun Rao and Aston Zhang and Aurelien Rodriguez and Austen Gregerson and Ava Spataru and Baptiste Roziere and Bethany Biron and Binh Tang and Bobbie Chern and Charlotte Caucheteux and Chaya Nayak and Chloe Bi and Chris Marra and Chris McConnell and Christian Keller and Christophe Touret and Chunyang Wu and Corinne Wong and Cristian Canton Ferrer and Cyrus Nikolaidis and Damien Allonsius and Daniel Song and Danielle Pintz and Danny Livshits and Danny Wyatt and David Esiobu and Dhruv Choudhary and Dhruv Mahajan and Diego Garcia-Olano and Diego Perino and Dieuwke Hupkes and Egor Lakomkin and Ehab AlBadawy and Elina Lobanova and Emily Dinan and Eric Michael Smith and Filip Radenovic and Francisco Guzmán and Frank Zhang and Gabriel Synnaeve and Gabrielle Lee and Georgia Lewis Anderson and Govind Thattai and Graeme Nail and Gregoire Mialon and Guan Pang and Guillem Cucurell and Hailey Nguyen and Hannah Korevaar and Hu Xu and Hugo Touvron and Iliyan Zarov and Imanol Arrieta Ibarra and Isabel Kloumann and Ishan Misra and Ivan Evtimov and Jack Zhang and Jade Copet and Jaewon Lee and Jan Geffert and Jana Vranes and Jason Park and Jay Mahadeokar and Jeet Shah and Jelmer van der Linde and Jennifer Billock and Jenny Hong and Jenya Lee and Jeremy Fu and Jianfeng Chi and Jianyu Huang and Jiawen Liu and Jie Wang and Jiecao Yu and Joanna Bitton and Joe Spisak and Jongsoo Park and Joseph Rocca and Joshua Johnstun and Joshua Saxe and Junteng Jia and Kalyan Vasuden Alwala and Karthik Prasad and Kartikeya Upasani and Kate Plawiak and Ke Li and Kenneth Heafield and Kevin Stone and Khalid El-Arini and Krithika Iyer and Kshitiz Malik and Kuenley Chiu and Kunal Bhalla and Kushal Lakhotia and Lauren Rantala-Yeary and Laurens van der Maaten and Lawrence Chen and Liang Tan and Liz Jenkins and Louis Martin and Lovish Madaan and Lubo Malo and Lukas Blecher and Lukas Landzaat and Luke de Oliveira and Madeline Muzzi and Mahesh Pasupuleti and Mannat Singh and Manohar Paluri and Marcin Kardas and Maria Tsimpoukelli and Mathew Oldham and Mathieu Rita and Maya Pavlova and Melanie Kambadur and Mike Lewis and Min Si and Mitesh Kumar Singh and Mona Hassan and Naman Goyal and Narjes Torabi and Nikolay Bashlykov and Nikolay Bogoychev and Niladri Chatterji and Ning Zhang and Olivier Duchenne and Onur Çelebi and Patrick Alrassy and Pengchuan Zhang and Pengwei Li and Petar Vasic and Peter Weng and Prajjwal Bhargava and Pratik Dubal and Praveen Krishnan and Punit Singh Koura and Puxin Xu and Qing He and Qingxiao Dong and Ragavan Srinivasan and Raj Ganapathy and Ramon Calderer and Ricardo Silveira Cabral and Robert Stojnic and Roberta Raileanu and Rohan Maheswari and Rohit Girdhar and Rohit Patel and Romain Sauvestre and Ronnie Polidoro and Roshan Sumbaly and Ross Taylor and Ruan Silva and Rui Hou and Rui Wang and Saghar Hosseini and Sahana Chennabasappa and Sanjay Singh and Sean Bell and Seohyun Sonia Kim and Sergey Edunov and Shaoliang Nie and Sharan Narang and Sharath Raparthy and Sheng Shen and Shengye Wan and Shruti Bhosale and Shun Zhang and Simon Vandenhende and Soumya Batra and Spencer Whitman and Sten Sootla and Stephane Collot and Suchin Gururangan and Sydney Borodinsky and Tamar Herman and Tara Fowler and Tarek Sheasha and Thomas Georgiou and Thomas Scialom and Tobias Speckbacher and Todor Mihaylov and Tong Xiao and Ujjwal Karn and Vedanuj Goswami and Vibhor Gupta and Vignesh Ramanathan and Viktor Kerkez and Vincent Gonguet and Virginie Do and Vish Vogeti and Vítor Albiero and Vladan Petrovic and Weiwei Chu and Wenhan Xiong and Wenyin Fu and Whitney Meers and Xavier Martinet and Xiaodong Wang and Xiaofang Wang and Xiaoqing Ellen Tan and Xide Xia and Xinfeng Xie and Xuchao Jia and Xuewei Wang and Yaelle Goldschlag and Yashesh Gaur and Yasmine Babaei and Yi Wen and Yiwen Song and Yuchen Zhang and Yue Li and Yuning Mao and Zacharie Delpierre Coudert and Zheng Yan and Zhengxing Chen and Zoe Papakipos and Aaditya Singh and Aayushi Srivastava and Abha Jain and Adam Kelsey and Adam Shajnfeld and Adithya Gangidi and Adolfo Victoria and Ahuva Goldstand and Ajay Menon and Ajay Sharma and Alex Boesenberg and Alexei Baevski and Allie Feinstein and Amanda Kallet and Amit Sangani and Amos Teo and Anam Yunus and Andrei Lupu and Andres Alvarado and Andrew Caples and Andrew Gu and Andrew Ho and Andrew Poulton and Andrew Ryan and Ankit Ramchandani and Annie Dong and Annie Franco and Anuj Goyal and Aparajita Saraf and Arkabandhu Chowdhury and Ashley Gabriel and Ashwin Bharambe and Assaf Eisenman and Azadeh Yazdan and Beau James and Ben Maurer and Benjamin Leonhardi and Bernie Huang and Beth Loyd and Beto De Paola and Bhargavi Paranjape and Bing Liu and Bo Wu and Boyu Ni and Braden Hancock and Bram Wasti and Brandon Spence and Brani Stojkovic and Brian Gamido and Britt Montalvo and Carl Parker and Carly Burton and Catalina Mejia and Ce Liu and Changhan Wang and Changkyu Kim and Chao Zhou and Chester Hu and Ching-Hsiang Chu and Chris Cai and Chris Tindal and Christoph Feichtenhofer and Cynthia Gao and Damon Civin and Dana Beaty and Daniel Kreymer and Daniel Li and David Adkins and David Xu and Davide Testuggine and Delia David and Devi Parikh and Diana Liskovich and Didem Foss and Dingkang Wang and Duc Le and Dustin Holland and Edward Dowling and Eissa Jamil and Elaine Montgomery and Eleonora Presani and Emily Hahn and Emily Wood and Eric-Tuan Le and Erik Brinkman and Esteban Arcaute and Evan Dunbar and Evan Smothers and Fei Sun and Felix Kreuk and Feng Tian and Filippos Kokkinos and Firat Ozgenel and Francesco Caggioni and Frank Kanayet and Frank Seide and Gabriela Medina Florez and Gabriella Schwarz and Gada Badeer and Georgia Swee and Gil Halpern and Grant Herman and Grigory Sizov and Guangyi and Zhang and Guna Lakshminarayanan and Hakan Inan and Hamid Shojanazeri and Han Zou and Hannah Wang and Hanwen Zha and Haroun Habeeb and Harrison Rudolph and Helen Suk and Henry Aspegren and Hunter Goldman and Hongyuan Zhan and Ibrahim Damlaj and Igor Molybog and Igor Tufanov and Ilias Leontiadis and Irina-Elena Veliche and Itai Gat and Jake Weissman and James Geboski and James Kohli and Janice Lam and Japhet Asher and Jean-Baptiste Gaya and Jeff Marcus and Jeff Tang and Jennifer Chan and Jenny Zhen and Jeremy Reizenstein and Jeremy Teboul and Jessica Zhong and Jian Jin and Jingyi Yang and Joe Cummings and Jon Carvill and Jon Shepard and Jonathan McPhie and Jonathan Torres and Josh Ginsburg and Junjie Wang and Kai Wu and Kam Hou U and Karan Saxena and Kartikay Khandelwal and Katayoun Zand and Kathy Matosich and Kaushik Veeraraghavan and Kelly Michelena and Keqian Li and Kiran Jagadeesh and Kun Huang and Kunal Chawla and Kyle Huang and Lailin Chen and Lakshya Garg and Lavender A and Leandro Silva and Lee Bell and Lei Zhang and Liangpeng Guo and Licheng Yu and Liron Moshkovich and Luca Wehrstedt and Madian Khabsa and Manav Avalani and Manish Bhatt and Martynas Mankus and Matan Hasson and Matthew Lennie and Matthias Reso and Maxim Groshev and Maxim Naumov and Maya Lathi and Meghan Keneally and Miao Liu and Michael L. Seltzer and Michal Valko and Michelle Restrepo and Mihir Patel and Mik Vyatskov and Mikayel Samvelyan and Mike Clark and Mike Macey and Mike Wang and Miquel Jubert Hermoso and Mo Metanat and Mohammad Rastegari and Munish Bansal and Nandhini Santhanam and Natascha Parks and Natasha White and Navyata Bawa and Nayan Singhal and Nick Egebo and Nicolas Usunier and Nikhil Mehta and Nikolay Pavlovich Laptev and Ning Dong and Norman Cheng and Oleg Chernoguz and Olivia Hart and Omkar Salpekar and Ozlem Kalinli and Parkin Kent and Parth Parekh and Paul Saab and Pavan Balaji and Pedro Rittner and Philip Bontrager and Pierre Roux and Piotr Dollar and Polina Zvyagina and Prashant Ratanchandani and Pritish Yuvraj and Qian Liang and Rachad Alao and Rachel Rodriguez and Rafi Ayub and Raghotham Murthy and Raghu Nayani and Rahul Mitra and Rangaprabhu Parthasarathy and Raymond Li and Rebekkah Hogan and Robin Battey and Rocky Wang and Russ Howes and Ruty Rinott and Sachin Mehta and Sachin Siby and Sai Jayesh Bondu and Samyak Datta and Sara Chugh and Sara Hunt and Sargun Dhillon and Sasha Sidorov and Satadru Pan and Saurabh Mahajan and Saurabh Verma and Seiji Yamamoto and Sharadh Ramaswamy and Shaun Lindsay and Shaun Lindsay and Sheng Feng and Shenghao Lin and Shengxin Cindy Zha and Shishir Patil and Shiva Shankar and Shuqiang Zhang and Shuqiang Zhang and Sinong Wang and Sneha Agarwal and Soji Sajuyigbe and Soumith Chintala and Stephanie Max and Stephen Chen and Steve Kehoe and Steve Satterfield and Sudarshan Govindaprasad and Sumit Gupta and Summer Deng and Sungmin Cho and Sunny Virk and Suraj Subramanian and Sy Choudhury and Sydney Goldman and Tal Remez and Tamar Glaser and Tamara Best and Thilo Koehler and Thomas Robinson and Tianhe Li and Tianjun Zhang and Tim Matthews and Timothy Chou and Tzook Shaked and Varun Vontimitta and Victoria Ajayi and Victoria Montanez and Vijai Mohan and Vinay Satish Kumar and Vishal Mangla and Vlad Ionescu and Vlad Poenaru and Vlad Tiberiu Mihailescu and Vladimir Ivanov and Wei Li and Wenchen Wang and Wenwen Jiang and Wes Bouaziz and Will Constable and Xiaocheng Tang and Xiaojian Wu and Xiaolan Wang and Xilun Wu and Xinbo Gao and Yaniv Kleinman and Yanjun Chen and Ye Hu and Ye Jia and Ye Qi and Yenda Li and Yilin Zhang and Ying Zhang and Yossi Adi and Youngjin Nam and Yu and Wang and Yu Zhao and Yuchen Hao and Yundi Qian and Yunlu Li and Yuzi He and Zach Rait and Zachary DeVito and Zef Rosnbrick and Zhaoduo Wen and Zhenyu Yang and Zhiwei Zhao and Zhiyu Ma},
      year={2024},
      eprint={2407.21783},
      archivePrefix={arXiv},
      primaryClass={cs.AI},
      url={https://arxiv.org/abs/2407.21783}, 
}

@misc{cobbe2021gsm8k,
      title={Training Verifiers to Solve Math Word Problems}, 
      author={Karl Cobbe and Vineet Kosaraju and Mohammad Bavarian and Mark Chen and Heewoo Jun and Lukasz Kaiser and Matthias Plappert and Jerry Tworek and Jacob Hilton and Reiichiro Nakano and Christopher Hesse and John Schulman},
      year={2021},
      eprint={2110.14168},
      archivePrefix={arXiv},
      primaryClass={cs.LG},
      url={https://arxiv.org/abs/2110.14168}, 
}

@inproceedings{
gandhi2024sos,
title={Stream of Search (SoS): Learning to Search in Language},
author={Kanishk Gandhi and Denise H J Lee and Gabriel Grand and Muxin Liu and Winson Cheng and Archit Sharma and Noah Goodman},
booktitle={First Conference on Language Modeling},
year={2024},
url={https://openreview.net/forum?id=2cop2jmQVL}
}

@inproceedings{he2024olympiad,
    title = "{O}lympiad{B}ench: A Challenging Benchmark for Promoting {AGI} with Olympiad-Level Bilingual Multimodal Scientific Problems",
    author = "He, Chaoqun  and
      Luo, Renjie  and
      Bai, Yuzhuo  and
      Hu, Shengding  and
      Thai, Zhen  and
      Shen, Junhao  and
      Hu, Jinyi  and
      Han, Xu  and
      Huang, Yujie  and
      Zhang, Yuxiang  and
      Liu, Jie  and
      Qi, Lei  and
      Liu, Zhiyuan  and
      Sun, Maosong",
    editor = "Ku, Lun-Wei  and
      Martins, Andre  and
      Srikumar, Vivek",
    booktitle = "Proceedings of the 62nd Annual Meeting of the Association for Computational Linguistics (Volume 1: Long Papers)",
    month = aug,
    year = "2024",
    address = "Bangkok, Thailand",
    publisher = "Association for Computational Linguistics",
    url = "https://aclanthology.org/2024.acl-long.211/",
    doi = "10.18653/v1/2024.acl-long.211",
    pages = "3828--3850"
}

@misc{austin2021mbpp,
      title={Program Synthesis with Large Language Models}, 
      author={Jacob Austin and Augustus Odena and Maxwell Nye and Maarten Bosma and Henryk Michalewski and David Dohan and Ellen Jiang and Carrie Cai and Michael Terry and Quoc Le and Charles Sutton},
      year={2021},
      eprint={2108.07732},
      archivePrefix={arXiv},
      primaryClass={cs.PL},
      url={https://arxiv.org/abs/2108.07732}, 
}

@inproceedings{mostafazadeh2016roc,
    title = "A Corpus and Cloze Evaluation for Deeper Understanding of Commonsense Stories",
    author = "Mostafazadeh, Nasrin  and
      Chambers, Nathanael  and
      He, Xiaodong  and
      Parikh, Devi  and
      Batra, Dhruv  and
      Vanderwende, Lucy  and
      Kohli, Pushmeet  and
      Allen, James",
    editor = "Knight, Kevin  and
      Nenkova, Ani  and
      Rambow, Owen",
    booktitle = "Proceedings of the 2016 Conference of the North {A}merican Chapter of the Association for Computational Linguistics: Human Language Technologies",
    month = jun,
    year = "2016",
    address = "San Diego, California",
    publisher = "Association for Computational Linguistics",
    url = "https://aclanthology.org/N16-1098/",
    doi = "10.18653/v1/N16-1098",
    pages = "839--849"
}

@misc{evalscope_2024,
    title={{EvalScope}: Evaluation Framework for Large Models},
    author={ModelScope Team},
    year={2024},
    url={https://github.com/modelscope/evalscope}
}

@article{sheng2024hybridflow,
  title   = {HybridFlow: A Flexible and Efficient RLHF Framework},
  author  = {Guangming Sheng and Chi Zhang and Zilingfeng Ye and Xibin Wu and Wang Zhang and Ru Zhang and Yanghua Peng and Haibin Lin and Chuan Wu},
  year    = {2024},
  journal = {arXiv preprint arXiv: 2409.19256}
}

@inproceedings{
liu2025understanding,
title={Understanding R1-Zero-Like Training: A Critical Perspective},
author={Zichen Liu and Changyu Chen and Wenjun Li and Penghui Qi and Tianyu Pang and Chao Du and Wee Sun Lee and Min Lin},
booktitle={Second Conference on Language Modeling},
year={2025},
url={https://openreview.net/forum?id=5PAF7PAY2Y}
}

@inproceedings{
gandhi2025cognitive,
title={Cognitive Behaviors that Enable Self-Improving Reasoners, or, Four Habits of Highly Effective {ST}aRs},
author={Kanishk Gandhi and Ayush K Chakravarthy and Anikait Singh and Nathan Lile and Noah Goodman},
booktitle={Second Conference on Language Modeling},
year={2025},
url={https://openreview.net/forum?id=QGJ9ttXLTy}
}

@inproceedings{
lambert2025tulu,
title={Tulu 3: Pushing Frontiers in Open Language Model Post-Training},
author={Nathan Lambert and Jacob Morrison and Valentina Pyatkin and Shengyi Huang and Hamish Ivison and Faeze Brahman and Lester James Validad Miranda and Alisa Liu and Nouha Dziri and Xinxi Lyu and Yuling Gu and Saumya Malik and Victoria Graf and Jena D. Hwang and Jiangjiang Yang and Ronan Le Bras and Oyvind Tafjord and Christopher Wilhelm and Luca Soldaini and Noah A. Smith and Yizhong Wang and Pradeep Dasigi and Hannaneh Hajishirzi},
booktitle={Second Conference on Language Modeling},
year={2025},
url={https://openreview.net/forum?id=i1uGbfHHpH}
}

@misc{openai2024openaio1card,
      title={OpenAI o1 System Card}, 
      author={OpenAI and : and Aaron Jaech and Adam Kalai and Adam Lerer and Adam Richardson and Ahmed El-Kishky and Aiden Low and Alec Helyar and Aleksander Madry and Alex Beutel and Alex Carney and Alex Iftimie and Alex Karpenko and Alex Tachard Passos and Alexander Neitz and Alexander Prokofiev and Alexander Wei and Allison Tam and Ally Bennett and Ananya Kumar and Andre Saraiva and Andrea Vallone and Andrew Duberstein and Andrew Kondrich and Andrey Mishchenko and Andy Applebaum and Angela Jiang and Ashvin Nair and Barret Zoph and Behrooz Ghorbani and Ben Rossen and Benjamin Sokolowsky and Boaz Barak and Bob McGrew and Borys Minaiev and Botao Hao and Bowen Baker and Brandon Houghton and Brandon McKinzie and Brydon Eastman and Camillo Lugaresi and Cary Bassin and Cary Hudson and Chak Ming Li and Charles de Bourcy and Chelsea Voss and Chen Shen and Chong Zhang and Chris Koch and Chris Orsinger and Christopher Hesse and Claudia Fischer and Clive Chan and Dan Roberts and Daniel Kappler and Daniel Levy and Daniel Selsam and David Dohan and David Farhi and David Mely and David Robinson and Dimitris Tsipras and Doug Li and Dragos Oprica and Eben Freeman and Eddie Zhang and Edmund Wong and Elizabeth Proehl and Enoch Cheung and Eric Mitchell and Eric Wallace and Erik Ritter and Evan Mays and Fan Wang and Felipe Petroski Such and Filippo Raso and Florencia Leoni and Foivos Tsimpourlas and Francis Song and Fred von Lohmann and Freddie Sulit and Geoff Salmon and Giambattista Parascandolo and Gildas Chabot and Grace Zhao and Greg Brockman and Guillaume Leclerc and Hadi Salman and Haiming Bao and Hao Sheng and Hart Andrin and Hessam Bagherinezhad and Hongyu Ren and Hunter Lightman and Hyung Won Chung and Ian Kivlichan and Ian O'Connell and Ian Osband and Ignasi Clavera Gilaberte and Ilge Akkaya and Ilya Kostrikov and Ilya Sutskever and Irina Kofman and Jakub Pachocki and James Lennon and Jason Wei and Jean Harb and Jerry Twore and Jiacheng Feng and Jiahui Yu and Jiayi Weng and Jie Tang and Jieqi Yu and Joaquin Quiñonero Candela and Joe Palermo and Joel Parish and Johannes Heidecke and John Hallman and John Rizzo and Jonathan Gordon and Jonathan Uesato and Jonathan Ward and Joost Huizinga and Julie Wang and Kai Chen and Kai Xiao and Karan Singhal and Karina Nguyen and Karl Cobbe and Katy Shi and Kayla Wood and Kendra Rimbach and Keren Gu-Lemberg and Kevin Liu and Kevin Lu and Kevin Stone and Kevin Yu and Lama Ahmad and Lauren Yang and Leo Liu and Leon Maksin and Leyton Ho and Liam Fedus and Lilian Weng and Linden Li and Lindsay McCallum and Lindsey Held and Lorenz Kuhn and Lukas Kondraciuk and Lukasz Kaiser and Luke Metz and Madelaine Boyd and Maja Trebacz and Manas Joglekar and Mark Chen and Marko Tintor and Mason Meyer and Matt Jones and Matt Kaufer and Max Schwarzer and Meghan Shah and Mehmet Yatbaz and Melody Y. Guan and Mengyuan Xu and Mengyuan Yan and Mia Glaese and Mianna Chen and Michael Lampe and Michael Malek and Michele Wang and Michelle Fradin and Mike McClay and Mikhail Pavlov and Miles Wang and Mingxuan Wang and Mira Murati and Mo Bavarian and Mostafa Rohaninejad and Nat McAleese and Neil Chowdhury and Neil Chowdhury and Nick Ryder and Nikolas Tezak and Noam Brown and Ofir Nachum and Oleg Boiko and Oleg Murk and Olivia Watkins and Patrick Chao and Paul Ashbourne and Pavel Izmailov and Peter Zhokhov and Rachel Dias and Rahul Arora and Randall Lin and Rapha Gontijo Lopes and Raz Gaon and Reah Miyara and Reimar Leike and Renny Hwang and Rhythm Garg and Robin Brown and Roshan James and Rui Shu and Ryan Cheu and Ryan Greene and Saachi Jain and Sam Altman and Sam Toizer and Sam Toyer and Samuel Miserendino and Sandhini Agarwal and Santiago Hernandez and Sasha Baker and Scott McKinney and Scottie Yan and Shengjia Zhao and Shengli Hu and Shibani Santurkar and Shraman Ray Chaudhuri and Shuyuan Zhang and Siyuan Fu and Spencer Papay and Steph Lin and Suchir Balaji and Suvansh Sanjeev and Szymon Sidor and Tal Broda and Aidan Clark and Tao Wang and Taylor Gordon and Ted Sanders and Tejal Patwardhan and Thibault Sottiaux and Thomas Degry and Thomas Dimson and Tianhao Zheng and Timur Garipov and Tom Stasi and Trapit Bansal and Trevor Creech and Troy Peterson and Tyna Eloundou and Valerie Qi and Vineet Kosaraju and Vinnie Monaco and Vitchyr Pong and Vlad Fomenko and Weiyi Zheng and Wenda Zhou and Wes McCabe and Wojciech Zaremba and Yann Dubois and Yinghai Lu and Yining Chen and Young Cha and Yu Bai and Yuchen He and Yuchen Zhang and Yunyun Wang and Zheng Shao and Zhuohan Li},
      year={2024},
      eprint={2412.16720},
      archivePrefix={arXiv},
      primaryClass={cs.AI},
      url={https://arxiv.org/abs/2412.16720}, 
}

@misc{kimi2025k15,
      title={Kimi k1.5: Scaling Reinforcement Learning with LLMs}, 
      author={Kimi Team and Angang Du and Bofei Gao and Bowei Xing and Changjiu Jiang and Cheng Chen and Cheng Li and Chenjun Xiao and Chenzhuang Du and Chonghua Liao and Chuning Tang and Congcong Wang and Dehao Zhang and Enming Yuan and Enzhe Lu and Fengxiang Tang and Flood Sung and Guangda Wei and Guokun Lai and Haiqing Guo and Han Zhu and Hao Ding and Hao Hu and Hao Yang and Hao Zhang and Haotian Yao and Haotian Zhao and Haoyu Lu and Haoze Li and Haozhen Yu and Hongcheng Gao and Huabin Zheng and Huan Yuan and Jia Chen and Jianhang Guo and Jianlin Su and Jianzhou Wang and Jie Zhao and Jin Zhang and Jingyuan Liu and Junjie Yan and Junyan Wu and Lidong Shi and Ling Ye and Longhui Yu and Mengnan Dong and Neo Zhang and Ningchen Ma and Qiwei Pan and Qucheng Gong and Shaowei Liu and Shengling Ma and Shupeng Wei and Sihan Cao and Siying Huang and Tao Jiang and Weihao Gao and Weimin Xiong and Weiran He and Weixiao Huang and Weixin Xu and Wenhao Wu and Wenyang He and Xianghui Wei and Xianqing Jia and Xingzhe Wu and Xinran Xu and Xinxing Zu and Xinyu Zhou and Xuehai Pan and Y. Charles and Yang Li and Yangyang Hu and Yangyang Liu and Yanru Chen and Yejie Wang and Yibo Liu and Yidao Qin and Yifeng Liu and Ying Yang and Yiping Bao and Yulun Du and Yuxin Wu and Yuzhi Wang and Zaida Zhou and Zhaoji Wang and Zhaowei Li and Zhen Zhu and Zheng Zhang and Zhexu Wang and Zhilin Yang and Zhiqi Huang and Zihao Huang and Ziyao Xu and Zonghan Yang and Zongyu Lin},
      year={2025},
      eprint={2501.12599},
      archivePrefix={arXiv},
      primaryClass={cs.AI},
      url={https://arxiv.org/abs/2501.12599}, 
}

@article{paszke2019pytorch,
  title={Pytorch: An imperative style, high-performance deep learning library},
  author={Paszke, Adam and Gross, Sam and Massa, Francisco and Lerer, Adam and Bradbury, James and Chanan, Gregory and Killeen, Trevor and Lin, Zeming and Gimelshein, Natalia and Antiga, Luca and others},
  journal={Advances in neural information processing systems},
  volume={32},
  year={2019}
}

@article{li2026learning,
  title={Learning robust reasoning through guided adversarial self-play},
  author={Li, Shuozhe and Tadiparthi, Vaishnav and Lee, Kwonjoon and Agarwal, Nakul and Mahjoub, Hossein Nourkhiz and Pari, Ehsan Moradi and Chen, Lizhang and Zhang, Amy and Leqi, Liu},
  journal={arXiv preprint arXiv:2602.00173},
  year={2026}
}

@article{liu2026palm,
  title={Palm: Progress-aware policy learning via affordance reasoning for long-horizon robotic manipulation},
  author={Liu, Yuanzhe and Zhu, Jingyuan and Mo, Yuchen and Li, Gen and Cao, Xu and Jin, Jin and Shen, Yifan and Li, Zhengyuan and Yu, Tianjiao and Yuan, Wenzhen and others},
  journal={arXiv preprint arXiv:2601.07060},
  year={2026}
}

@article{shen2026fine,
  title={Fine-grained preference optimization improves spatial reasoning in vlms},
  author={Shen, Yifan and Liu, Yuanzhe and Zhu, Jingyuan and Cao, Xu and Zhang, Xiaofeng and He, Yixiao and Ye, Wenming and Rehg, James and Lourentzou, Ismini},
  journal={Advances in Neural Information Processing Systems},
  volume={38},
  pages={17929--17960},
  year={2026}
}

@article{li2026toward,
  title={Toward cognitive supersensing in multimodal large language model},
  author={Li, Boyi and Shen, Yifan and Liu, Yuanzhe and Xu, Yifan and Liu, Jiateng and Li, Xinzhuo and Li, Zhengyuan and Zhu, Jingyuan and Zhong, Yunhan and Lan, Fangzhou and others},
  journal={arXiv preprint arXiv:2602.01541},
  year={2026}
}

@article{yu2025core3d,
  title={Core3d: Collaborative reasoning as a foundation for 3d intelligence},
  author={Yu, Tianjiao and Li, Xinzhuo and Shen, Yifan and Liu, Yuanzhe and Lourentzou, Ismini},
  journal={arXiv preprint arXiv:2512.12768},
  year={2025}
}

@article{luo2026learning,
  title={Learning from the Self-future: On-policy Self-distillation for dLLMs},
  author={Luo, Yifu and Chen, Zeyu and Wang, Haoyu and Hu, Xinhao and Zhang, Yuxuan and Sha, Zhizhou and Liu, Shiwei},
  journal={arXiv preprint arXiv:2606.18195},
  year={2026}
}

@article{luo2026reinforcement,
  title={Reinforcement learning meets masked generative models: Mask-grpo for text-to-image generation},
  author={Luo, Yifu and Hu, Xinhao and Fan, Keyu and Sun, Haoyuan and Chen, Zeyu and Xia, Bo and Zhang, Tiantian and Chang, Yongzhe and Wang, Xueqian},
  journal={Advances in Neural Information Processing Systems},
  volume={38},
  pages={108460--108485},
  year={2026}
}

@InProceedings{qiu2026evolutionstrategies,
  title = 	 {Evolution Strategies at Scale: {LLM} Fine-Tuning Beyond Reinforcement Learning},
  author =       {Qiu, Xin and Gan, Yulu and Hayes, Conor F. and Liang, Qiyao and Xu, Yinggan and Dailey, Roberto and Meyerson, Elliot and Hodjat, Babak and Miikkulainen, Risto},
  booktitle = 	 {Proceedings of the 43rd International Conference on Machine Learning},
  pages = 	 {101987--102015},
  year = 	 {2026},
  editor = 	 {Zhang, Tong and Dudik, Miroslav and Jaggi, Martin and Agarwal, Alekh and Li, Sharon and Schuurmans, Dale and Zhu, Jerry and Berkenkamp, Felix and Dong, Hanze and Bietti, Alberto},
  volume = 	 {306},
  series = 	 {Proceedings of Machine Learning Research},
  month = 	 {06--11 Jul},
  publisher =    {PMLR},
  url = 	 {https://proceedings.mlr.press/v306/qiu26f.html}
}

@misc{zhang2026datasetmismatchmattersgroup,
      title={Dataset Mismatch Matters in Group Relative Policy Optimization for Reinforcement Learning from Verifiable Rewards}, 
      author={Zhen-Yu Zhang and Jiandong Zhang and Huaxiu Yao and Gang Niu and Masashi Sugiyama},
      year={2026},
      eprint={2611.00586},
      archivePrefix={arXiv},
      primaryClass={cs.LG},
      url={https://arxiv.org/abs/2611.00586}, 
}
}

\newpage
\appendix



\section{Additional Empirical Results}
\seclabel{add_results}

This section reports four additional empirical studies that complement the main results: a component ablation of CoRP on representative cells, a comparison to naive and merging baselines on the same cells, a sensitivity sweep of the compatibility weights, and an analysis of the effect of perturbation population size.

\subsection{Component Ablation}
\seclabel{appx:component-ablation}

We isolate the contribution of each CoRP component on four cells covering two model scales and two task types. Five variants share the rewarded population and the support folds with full CoRP, and differ only in which component is removed. The variants are reward only ($\gamma_a{=}\gamma_d{=}0$, a single reward-weighted average), reward+alignment ($\gamma_d{=}0$), reward+dispersion ($\gamma_a{=}0$), no gate (deploys the best-on-$A$ candidate without the validation of \secref{crossfit}), and no iteration (skips the recentering of \secref{refine}). Table~\ref{tab:component} reports test accuracy and the change relative to the base model.

\begin{table}[ht]
\centering
\footnotesize
\caption{Component ablation on Qwen2.5-3B-Instruct and OLMo3-7B-Instruct, across GSM8K and ROCStories. Each entry reports test accuracy and, in parentheses, the change relative to the base model on the same cell. Full CoRP corresponds to the configuration of \Cref{tab:CoRP_filtered_comparison}.}
\label{tab:component}
\begin{tabular}{lcccc}
\toprule
 & \multicolumn{2}{c}{GSM8K} & \multicolumn{2}{c}{ROCStories} \\
\cmidrule(lr){2-3}\cmidrule(lr){4-5}
Method & Acc. & $\Delta$ & Acc. & $\Delta$ \\
\midrule
\rowcolor[rgb]{0.95,0.95,0.95}\multicolumn{5}{l}{\textit{Qwen2.5-3B-Instruct}} \\
Base                       & $79.81$ & --        & $54.73$ & --        \\
\quad reward only          & $80.97$ & $+1.16$   & $54.70$ & $-0.03$   \\
\quad reward+alignment     & $80.89$ & $+1.08$   & $53.97$ & $-0.76$   \\
\quad reward+dispersion    & $79.76$ & $-0.05$   & $54.66$ & $-0.07$   \\
\quad no gate              & $80.67$ & $+0.86$   & $54.59$ & $-0.14$   \\
\quad no iteration         & $\mathbf{82.31}$ & $\mathbf{+2.50}$ & $56.71$ & $+1.98$   \\
\quad \textbf{Full CoRP}   & $\mathbf{82.31}$ & $\mathbf{+2.50}$ & $\mathbf{59.11}$ & $\mathbf{+4.38}$ \\
\midrule
\rowcolor[rgb]{0.95,0.95,0.95}\multicolumn{5}{l}{\textit{OLMo3-7B-Instruct}} \\
Base                       & $82.92$ & --        & $64.04$ & --        \\
\quad reward only          & $84.16$ & $+1.24$   & $63.31$ & $-0.73$   \\
\quad reward+alignment     & $84.69$ & $+1.77$   & $62.82$ & $-1.22$   \\
\quad reward+dispersion    & $84.57$ & $+1.65$   & $63.40$ & $-0.64$   \\
\quad no gate              & $\mathbf{85.63}$ & $\mathbf{+2.71}$  & $63.66$ & $-0.38$  \\
\quad no iteration         & $84.82$ & $+1.90$   & $64.61$ & $+0.57$   \\
\quad \textbf{Full CoRP}   & $84.82$ & $+1.90$   & $\mathbf{64.81}$ & $\mathbf{+0.77}$ \\
\bottomrule
\end{tabular}
\end{table}

Each component contributes. The iteration step helps most on Qwen2.5-3B / ROCStories, where Full CoRP improves over the no-iteration variant by 2.40 points. On the other three cells, the accepted first-pass update already captures most of the measured gain. The held-out gate plays a complementary role by rejecting harmful candidates on cells where rewarded perturbations are less aligned. Dispersion is the weakest stand-alone choice, which matches its role in Eq.~9: it suppresses incompatible mass against a direction the alignment term has already proposed, and is informative only when paired with that direction.

\subsection{Naive Baselines}
\seclabel{appx:naive-baselines}

The reproducible low-rank structure of \secref{sec:characterization} raises a natural concern: simpler operators that read this structure directly might already suffice. Existing parameter-space merging tools, designed for fine-tuned checkpoints, raise a separate concern about whether they transfer to rewarded perturbations. We compare CoRP to three alternatives that probe both concerns on the same four cells used in \secref{appx:component-ablation}.

\textit{Reward-weighted averaging} is the operator obtained by setting $\gamma_a{=}\gamma_d{=}0$ in Eq.~\ref{eq:w2}, identical to the reward-only variant of \Cref{tab:component}. \textit{Top-$r$ subspace projection} estimates the top-$r$ principal subspace of the elite perturbations with $r{=}8$ and deploys $\theta_0 + \eta\, \Pi_U \bar{\Delta}$, where $\Pi_U$ projects onto the estimated subspace. \textit{Sparse merging in the spirit of TIES}~\citep{yadav2023ties} selects the top half of the rewarded population, sparsifies each perturbation by retaining the top-density fraction of coordinates by magnitude, and merges the sparsified vectors. We sweep the sparsity density and deployment scale on the probe fold for each cell. All three baselines share the rewarded population and the support folds with full CoRP. \Cref{tab:naive} reports the result.

\begin{table}[ht]
\centering
\footnotesize
\caption{Naive and merging baselines on Qwen2.5-3B-Instruct and OLMo3-7B-Instruct, across GSM8K and ROCStories. Each entry reports test accuracy and, in parentheses, the change relative to the base model on the same cell.}
\label{tab:naive}
\begin{tabular}{lcccc}
\toprule
 & \multicolumn{2}{c}{GSM8K} & \multicolumn{2}{c}{ROCStories} \\
\cmidrule(lr){2-3}\cmidrule(lr){4-5}
Method & Acc. & $\Delta$ & Acc. & $\Delta$ \\
\midrule
\rowcolor[rgb]{0.95,0.95,0.95}\multicolumn{5}{l}{\textit{Qwen2.5-3B-Instruct}} \\
Base                              & $79.81$ & --        & $54.73$ & --        \\
\quad reward-weighted avg.        & $80.97$ & $+1.16$   & $54.70$ & $-0.03$   \\
\quad top-$r$ projection          & $80.59$ & $+0.78$   & $54.59$ & $-0.14$   \\
\quad sparse merging (TIES-style) & $80.42$ & $+0.61$   & $53.91$ & $-0.82$   \\
\quad \textbf{Full CoRP}          & $\mathbf{82.31}$ & $\mathbf{+2.50}$ & $\mathbf{59.11}$ & $\mathbf{+4.38}$ \\
\midrule
\rowcolor[rgb]{0.95,0.95,0.95}\multicolumn{5}{l}{\textit{OLMo3-7B-Instruct}} \\
Base                              & $82.92$ & --        & $64.04$ & --        \\
\quad reward-weighted avg.        & $84.16$ & $+1.24$   & $63.31$ & $-0.73$   \\
\quad top-$r$ projection          & $83.85$ & $+0.93$   & $63.85$ & $-0.19$   \\
\quad sparse merging (TIES-style) & $83.32$ & $+0.40$   & $62.32$ & $-1.72$   \\
\quad \textbf{Full CoRP}          & $\mathbf{84.82}$ & $\mathbf{+1.90}$ & $\mathbf{64.81}$ & $\mathbf{+0.77}$ \\
\bottomrule
\end{tabular}
\end{table}

Reward-weighted averaging is the most competitive of the three baselines, but still trails CoRP across the four cells. Reward identifies useful perturbations, but compatibility against the proposal determines which of those perturbations can share a single deployable update.

Top-$r$ projection rarely moves beyond the base. The shared low-rank structure of \secref{sec:characterization} is a necessary condition for consolidation to be possible, but the population-level subspace estimate discards the per-perturbation reward signal that the operator of \secref{operator} actually uses, and the estimate is itself noisy at the population sizes we use.

Sparse merging in the spirit of TIES underperforms the base on two of four cells. Standard merging operators were designed for independently fine-tuned checkpoints whose disagreements carry meaningful sign and magnitude structure. The same operators applied to random rewarded perturbations have no comparable signal to read, since coordinate sign is essentially random and magnitude varies with the noise scale rather than with task relevance.
\subsection{Sensitivity to Compatibility Weights}
\seclabel{appx:gamma-sweep}

\begin{figure}[ht]\centering
\includegraphics[width=0.5\textwidth]{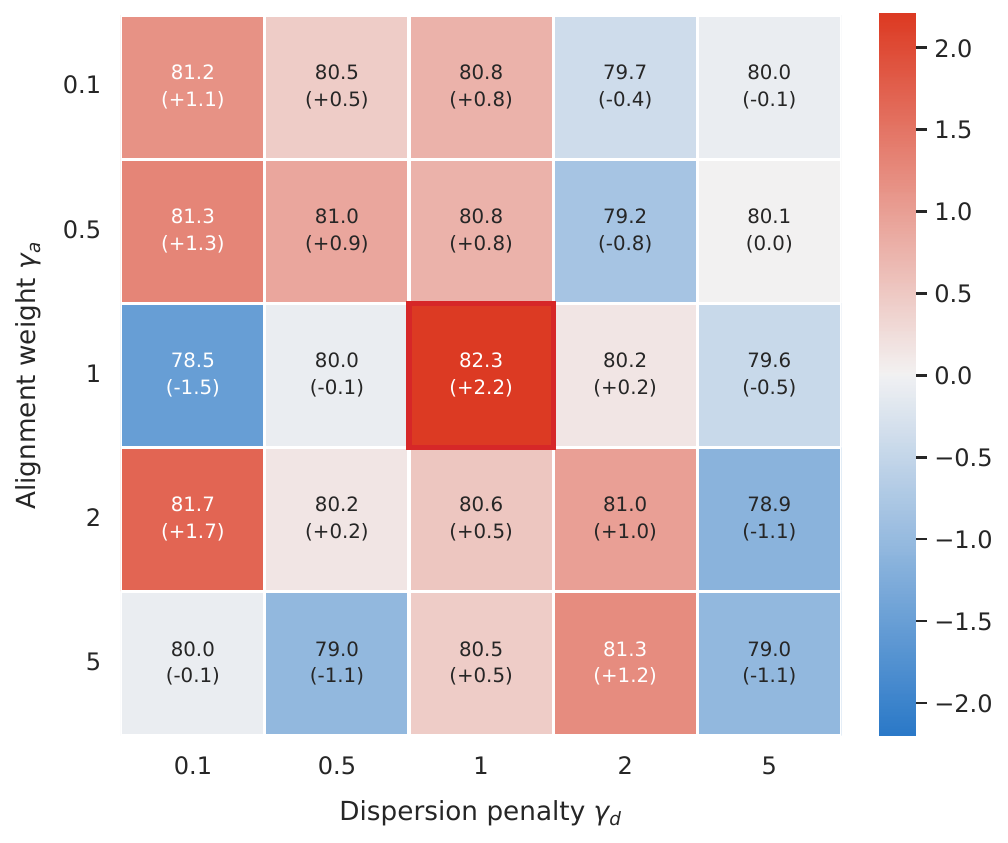}
\caption{Test accuracy on Qwen2.5-3B-Instruct / GSM8K under the first-pass CoRP operator, swept over the alignment weight $\gamma_a$ and dispersion penalty $\gamma_d$. Each cell reports accuracy and, in parentheses, the change relative to the base model (base $80.06$). The default $\gamma_a{=}\gamma_d{=}1$ is the best configuration on this sweep.}
\label{fig:gamma_sweep}
\end{figure}

The default configuration of CoRP fixes the alignment and dispersion weights at $\gamma_a{=}\gamma_d{=}1$ across all $25$ model-task pairs. To check that this choice is not fragile, we sweep both weights over $\{0.1, 0.5, 1, 2, 5\}$ on Qwen2.5-3B-Instruct / GSM8K, holding the rewarded population, the support folds, and the candidate grid over $(q, \beta, \alpha)$ identical to the main run. Each of the $25$ configurations rebuilds the consolidated update of \secref{operator} from the same perturbations and is evaluated on the GSM8K test set. The sweep uses the first-pass operator only, without the held-out gate of \secref{crossfit} or the iteration of \secref{refine}, so that the variation in the table reflects only the choice of $(\gamma_a, \gamma_d)$.

The default $(1, 1)$ is the strongest configuration on the sweep, and the central region $[0.5,2]^2$ remains mostly positive, with small negative outliers at a few asymmetric settings. The accuracy surface drops at the corners where one weight saturates the exponent in Eq.~\ref{eq:w2} and the other becomes negligible, which is the regime in which compatibility scoring effectively reduces to one of the two terms. Within the operating range we use, the operator is not sensitive to the precise value of either weight.

\subsection{Effect of Perturbation Population Size}
\seclabel{appx:n-scaling}

\begin{figure}[ht]
\centering
\includegraphics[width=0.9\textwidth]{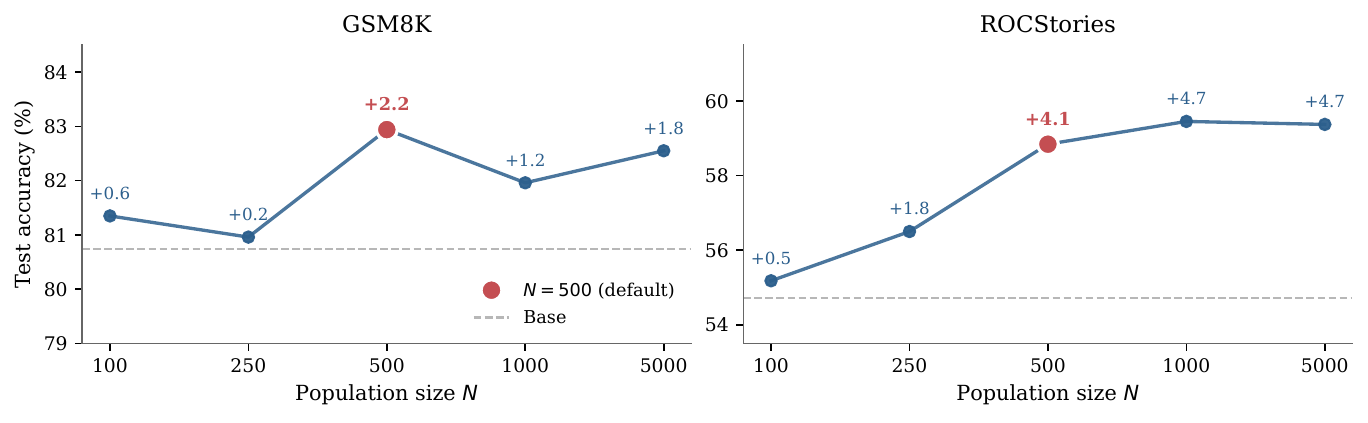}
\caption{CoRP test accuracy on Qwen2.5-3B-Instruct as a function of perturbation population size $N$. The dashed line marks the base model and the parenthesized number annotates the gain over base. The default $N{=}500$ used in the main results is highlighted in red. The sweep uses the first-pass operator only.}
\label{fig:n-scaling}
\end{figure}

CoRP uses $N{=}500$ rewarded perturbations throughout the main results, one tenth of the $5000$ used by RandOpt. To check that this choice is not arbitrary, we sweep $N \in \{100, 250, 500, 1000, 5000\}$ on Qwen2.5-3B-Instruct, holding the noise-scale mixture, the support folds, and the candidate grid identical to the main run. Each configuration runs the first-pass operator of \secref{operator}, without the iteration of \secref{refine}, so that the variation in \Cref{fig:n-scaling} reflects only the population size.

The marginal return from a larger population diminishes well before $N{=}5000$ on both tasks. On ROCStories, the gain plateaus between $N{=}1000$ and $N{=}5000$. On GSM8K, the gain peaks at $N=500$ in this sweep and remains positive but non-monotone for larger populations. The default $N{=}500$ recovers a substantial fraction of the maximum gain on both tasks at one tenth of the population size used by RandOpt.


\section{Additional Analysis on Consolidation Behavior}
\label{sec:appendix-analysis}

We provide two further analyses of consolidation behavior: how CoRP composes its test-set outcomes relative to ensemble baselines, and how strict vs.\ relaxed evaluation interacts with each method.

\subsection{Error-Composition Decomposition}

\begin{figure}[ht]
    \centering
    \includegraphics[width=0.55\textwidth]{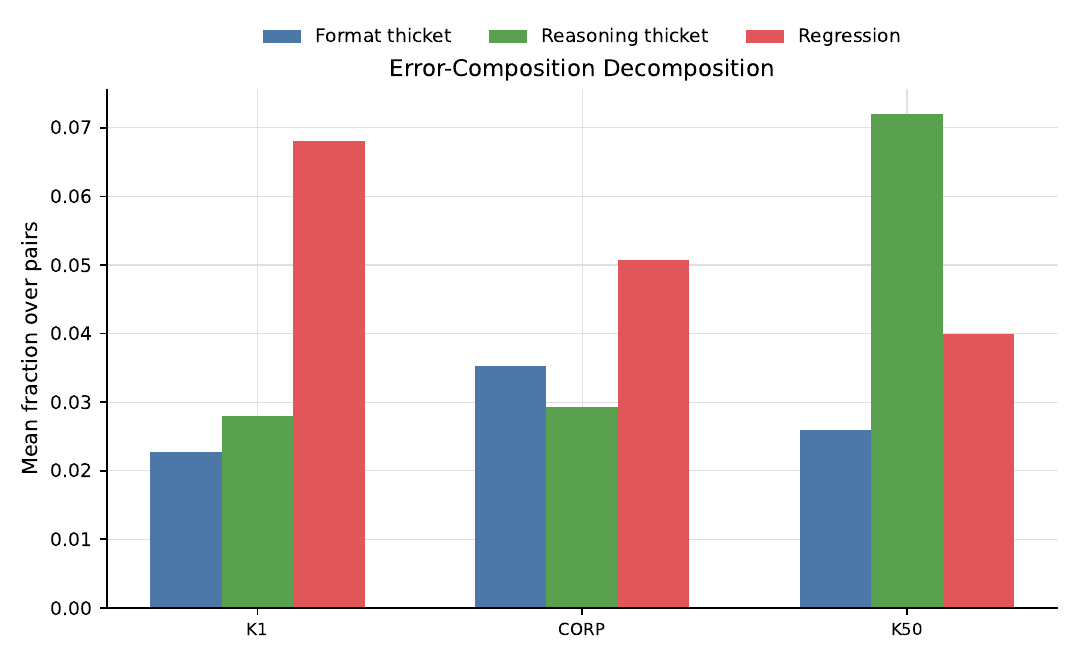}
    \caption{Mean bucket fractions across model-task pairs for RandOpt single-inference ($K{=}1$), CoRP, and 50-pass majority vote ($K{=}50$).}
    \label{fig:appendix-buckets}
\end{figure}

\Cref{fig:appendix-buckets} reveals a structured trade-off across the three deployment modes. Relative to $K{=}1$, CoRP reduces the average regression fraction while preserving a comparable reasoning-thicket fraction, indicating that compatibility-aware reweighting and held-out gating mitigate harmful updates without sacrificing reasoning gains. The 50-pass vote attains the lowest regression level by retaining inference-time diversity, but at substantially higher inference cost. CoRP therefore lands between the two on the regression-vs-inference-cost frontier, achieving the regression reduction of an ensemble at the inference cost of a single model.

\subsection{Robustness to Strict Evaluation}

\begin{figure}[ht]
    \centering
    \includegraphics[width=0.55\textwidth]{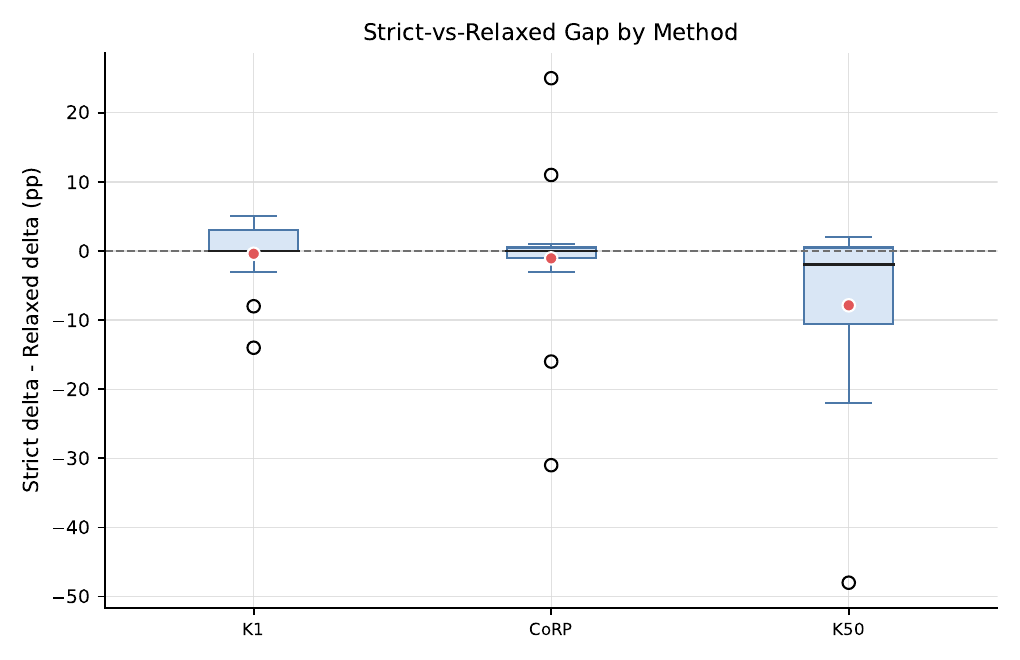}
    \caption{Distribution of strict-minus-relaxed deltas (percentage points) across model-task pairs. More negative values indicate that a method's gain shrinks more under strict parsing relative to relaxed evaluation.}
    \label{fig:appendix-strict-relaxed}
\end{figure}

\Cref{fig:appendix-strict-relaxed} reports how each method's gain holds up under the strict task evaluator used throughout the paper. The 50-pass vote has the most negative average strict-minus-relaxed gap, suggesting that part of its reward-level gain is attenuated under strict parsing. CoRP remains closer to the single-inference baseline on this axis, indicating that consolidation does not systematically amplify strict-format fragility. For deployment with strict extractors this distinction is practically important, because the gap between an ensemble's relaxed score and its deployed strict score is exactly what a single deployable model has to recover.

\section{Implementation Details}
\seclabel{reproducibility}

We implement CoRP in PyTorch~\citep{paszke2019pytorch} and all experiments were conducted on 4 NVIDIA A100 GPUs with 80GB memory.

\subsection{Support Splits and Model Selection}
For each model-task pair, we partition the support set into three disjoint folds: a construction fold $A$, a validation fold $B$, and a probe fold $P$. Fold $A$ scores perturbations and generates CoRP candidates. Fold $B$ gates candidates and selects among the $(q,\beta)$ grid. Fold $P$ is used only after a candidate passes the gate, to choose the final step-size multiplier $\alpha$ from the fixed grid in~\Cref{tab:hyperparams}. The test set plays no role in selecting $q$, $\beta$, or $\alpha$, nor in deciding whether to accept an update.

This design separates proposal construction from validation. If no candidate achieves a positive construction score and a positive lower-confidence-bound improvement on fold $B$ (defined in \secref{appx:operational-defs}), CoRP abstains and returns the base model. If a candidate passes the gate but no step-size multiplier passes the probe check on $P$, CoRP also abstains.

\subsection{Baseline Protocols}
Post-training helps adapt LLMs to a broad range of tasks and domains~\citep{ouyang2022rlhf, shao2024grpo, agarwal2024onpolicykd, liu2026palm,li2026toward,shen2026fine,yu2025core3d,luo2026learning,luo2026reinforcement,zhang2026datasetmismatchmattersgroup}. We compare CoRP against PPO, GRPO, RandOpt $(K=1)$, and RandOpt $(K=50)$ under the protocol of~\citet{gan2026neuralthickets}. RandOpt $(K=1)$ selects the single highest-reward perturbation from the sampled population; RandOpt $(K=50)$ ensembles the top 50 perturbations by majority vote at inference. RandOpt uses $N=5000$ perturbations and CoRP uses $N=500$. For PPO and GRPO, we follow the matched-compute settings reported by~\citet{gan2026neuralthickets}: total training FLOPs are matched to those consumed by sampling and scoring the RandOpt $K{=}50$ population, accounting for actor forward and backward passes, KL evaluation against the reference model, and rollout sequence lengths. We use the same models, tasks, prompts, reward functions, and evaluation scripts throughout. No baseline hyperparameters are tuned on the test set.

\subsection{Statistical Reporting}
The main result table reports mean$\pm$standard deviation over $R=3$ independent runs. For CoRP and RandOpt, each run resamples the perturbation population and repeats the full selection procedure; for PPO and GRPO, each run uses an independent training seed under the same matched-compute setting. Error bars summarize run-to-run variability and are not intended as formal pairwise hypothesis tests.

For the split-half diagnostics of \secref{sec:characterization}, we repeat the random partition of the top-$M$ rewarded perturbations $20$ times and report a nonparametric $95\%$ lower confidence bound via bootstrapping, taking the $5$th percentile of the bootstrap distribution of the mean. This procedure applies to both the mean-consensus statistic and the subspace-excess statistic, and is distinct from the gate-and-probe lower confidence bounds defined in \secref{appx:operational-defs}.

\subsection{Hyperparameter Settings}
\seclabel{hyperparams}

Table~\ref{tab:hyperparams} lists the CoRP hyperparameters. Values are fixed across all $25$ model-task pairs unless noted.

\begin{table}[htbp]
\centering
\footnotesize
\caption{The hyperparameter configuration of our experiments. Values are fixed across model-task pairs unless noted.}
\label{tab:hyperparams}
\renewcommand{\arraystretch}{1.1}
\begin{tabular}{lll}
\toprule
Symbol & Value & Notes \\
\midrule
\rowcolor[rgb]{0.95,0.95,0.95}\multicolumn{3}{l}{\emph{(A) Sampling}} \\
$N$ & 500 & rewarded perturbations per pair \\
$\Sigma$ & $\{5{\times}10^{-4},\, 1{\times}10^{-3},\, 2{\times}10^{-3}\}$ & noise-scale mixture, uniform per perturbation \\
$|A|, |B|, |P|$ & $75, 75, 50$ & first 200 training examples per task \\
\midrule
\rowcolor[rgb]{0.95,0.95,0.95}\multicolumn{3}{l}{\emph{(B) Candidate and gate}} \\
$q$ & $\{0.5,\, 0.7,\, 0.9\}$ & elite quantile, selected on $B$ \\
$\beta$ & $\{0.5,\, 1,\, 2,\, 5,\, 10,\, 20,\, 50\}$ & inverse temperature, selected on $B$ \\
$\gamma_a,\, \gamma_d$ & $1,\, 1$ & alignment and dispersion weights \\
$\lambda$ & $2.0$ & regression penalty in the constructive score \\
$z_{\mathrm{LCB}}$ & $1.645$ & one-sided 95\% normal-approximation LCB \\
$\{\alpha_j\}$ & $\{0.5,\, 1,\, 2,\, 4,\, 8,\, 16\}$ & probe multiplier grid, selected on $P$ \\
\midrule
\rowcolor[rgb]{0.95,0.95,0.95}\multicolumn{3}{l}{\emph{(C) Iteration}} \\
$N_{\mathrm{loc}}$ & $100$ & local perturbations per iteration \\
$r$ & $8$ & truncation rank of $\Sigma_t$ \\
$\lambda_{\mathrm{iso}}$ & $0.5$ & isotropic floor, clipped to $[0.05, 0.95]$ \\
\bottomrule
\end{tabular}
\end{table}

\subsection{Operational Definitions}
\seclabel{appx:operational-defs}

We give precise definitions for three quantities referenced in the main text: the coordinate sketch used by the split-half diagnostic, the proposal covariance $\Sigma_t$ used by the iteration step, and the lower confidence bounds used by the validation gate and probe.

\paragraph{Coordinate sketch.}
The split-half diagnostic of \secref{sec:characterization} reads each statistic from a fixed low-dimensional sketch rather than from $\mathbb{R}^d$ directly. For each perturbable parameter tensor in the model, we retain a small fixed number of coordinates from the flattened perturbation, and concatenate the retained coordinates across all tensors to form $z_i \in \mathbb{R}^m$ with $m$ on the order of a few thousand. The same coordinate selection is shared across all candidates in the population and across all $20$ random splits. We use the same sketch when comparing reward-weighted means and when computing principal subspaces, normalizing $z_i$ by its noise scale $\sigma_i$ to compare directions rather than magnitudes.

\paragraph{Proposal covariance.}
The iteration step of \secref{refine} draws local perturbations $\Delta_j^{\mathrm{loc}} \sim \mathcal{N}(0, \rho_t^2 \Sigma_t)$, where the covariance $\Sigma_t$ combines a low-rank component fit to the previous accepted population with an isotropic floor:
\begin{equation}
\Sigma_t \;=\; \lambda_{\mathrm{iso}}\, I_d \;+\; (1-\lambda_{\mathrm{iso}})\, U_t \Lambda_t U_t^\top,
\label{eq:sigmat-appendix}
\end{equation}
where $(U_t, \Lambda_t)$ is the rank-$r$ truncated eigendecomposition of the compatibility-weighted second-moment matrix of the previous round's elite perturbations, and the rank $r$ and floor $\lambda_{\mathrm{iso}}$ are reported in \Cref{tab:hyperparams}. The covariance shapes the local proposal distribution only and does not appear in the deployed update.

\paragraph{Lower confidence bounds.}
The gate and probe of \secref{crossfit} commit a candidate only when its estimated accuracy change has a positive lower confidence bound on the corresponding fold. For each candidate, we compute paired per-example correctness differences against the base model and use
\begin{equation}
\widehat{\mathrm{LCB}}
=
\widehat{\Delta\mathrm{Acc}}
-
z_{\mathrm{LCB}}\,
\widehat{\mathrm{SE}}(\Delta\mathrm{Acc}).
\end{equation}
where $z_{\mathrm{LCB}}$ is given in \Cref{tab:hyperparams}. This bound is computed separately for each candidate and fold, and is distinct from the bootstrap procedure used for the split-half diagnostics in \secref{sec:characterization}.

\subsection{Prompts}

Following~\citet{gan2026neuralthickets}, we set up the prompts for different datasets in our experiments following EvalScope~\citep{evalscope_2024} and Verl~\citep{sheng2024hybridflow}.

\begin{tcolorbox}[
  width=\textwidth,
  colframe=gray!50!black,
  colback=gray!5!white,
  title={\texttt{Countdown}},
]
\small
\textbf{Your Task}\par
Using the numbers \verb|{numbers}|, create an equation that equals \verb|{target}|.\par
\medskip
\textbf{Instructions}\par
You can use basic arithmetic operations (\verb|+|, \verb|-|, \verb|*|, \verb|/|), and each number can only be used once.\par
Show your work in \verb|<work>| \ldots{} \verb|</work>| tags.\par
Return the final answer in \verb|<answer>| \ldots{} \verb|</answer>| tags.\par
For example: \verb|<answer> (1 + 2) / 3 </answer>|.
\end{tcolorbox}

\medskip

\begin{tcolorbox}[
  width=\textwidth,
  colframe=gray!50!black,
  colback=gray!5!white,
  title={\texttt{GSM8K / OlympiadBench}},
]
\small
\textbf{Your Task}\par
Solve the following problem: \verb|{question}|.\par
\medskip
\textbf{Instructions}\par
Let's think step by step and output the final answer after \verb|####|.
\end{tcolorbox}

\medskip

\begin{tcolorbox}[
  width=\textwidth,
  colframe=gray!50!black,
  colback=gray!5!white,
  title={\texttt{ROCStories}},
]
\small
\textbf{Your Task}\par
Below are 5 sentences from a story, but they are in the wrong order.\par
Please arrange them in the correct chronological order.\par
\medskip
\textbf{Story}\par
Title: \verb|{title}|\par
Sentence A: \verb|{sentence1}|\par
Sentence B: \verb|{sentence2}|\par
Sentence C: \verb|{sentence3}|\par
Sentence D: \verb|{sentence4}|\par
Sentence E: \verb|{sentence5}|\par
\medskip
\textbf{Instructions}\par
Output the correct order as comma-separated letters (e.g., \verb|B,A,D,E,C|).\par
Only output the letters, nothing else.
\end{tcolorbox}

\medskip

\begin{tcolorbox}[
  width=\textwidth,
  colframe=gray!50!black,
  colback=gray!5!white,
  title={\texttt{MBPP}},
]
\small
\textbf{Your Task}\par
You are an expert Python programmer, and here is your task: \verb|{question}|\par
\medskip
\textbf{Instructions}\par
Your code should pass these tests: \verb|{tests}|.
\end{tcolorbox}

\section{Additional Acknowledgment}
The authors acknowledge the use of ChatGPT exclusively to refine the text in the final manuscript.


\newpage

\end{document}